%% file: arxiv.tex
\documentclass[10pt,twocolumn,letterpaper]{article}

\usepackage[pagenumbers]{cvpr} 
\usepackage{multirow}
\usepackage{bbding}
\usepackage{setspace}


\definecolor{cvprblue}{rgb}{0.21,0.49,0.74}
\usepackage[pagebackref,breaklinks,colorlinks,allcolors=cvprblue]{hyperref}

\usepackage{hyperref}
\usepackage{url}
\usepackage{footnote}
\hypersetup{
colorlinks,
urlcolor={red}
}

\title{FaithDiff: Unleashing Diffusion Priors for Faithful Image Super-resolution}

\author{Junyang Chen \quad Jinshan Pan \quad Jiangxin Dong$^{\dagger}$\\
School of Computer Science and Engineering, Nanjing University of Science and Technology\\
{\tt \url{https://jychen9811.github.io/FaithDiff_page/}}
}

\begin{document}
\maketitle

\def\thefootnote{$\dagger$}
\footnotetext{Corresponding author}

\begin{abstract}
Faithful image super-resolution (SR) not only needs to recover images that appear realistic, similar to image generation tasks, but also requires that the restored images maintain fidelity and structural consistency with the input.
To this end, we propose a simple and effective method, named FaithDiff, to fully harness the impressive power of latent diffusion models (LDMs) for faithful image SR.
In contrast to existing diffusion-based SR methods that freeze the diffusion model pre-trained on high-quality images, we propose to unleash the diffusion prior to identify useful information and recover faithful structures.
As there exists a significant gap between the features of degraded inputs and the noisy latent from the diffusion model, we then develop an effective alignment module to explore useful features from degraded inputs to align well with the diffusion process.
Considering the indispensable roles and interplay of the encoder and diffusion model in LDMs, we jointly fine-tune them in a unified optimization framework, facilitating the encoder to extract useful features that coincide with diffusion process.
Extensive experimental results demonstrate that FaithDiff outperforms state-of-the-art methods, providing high-quality and faithful SR results.

\end{abstract}


\section{Introduction}
\label{sec:intro}

Image super-resolution (SR) aims to recover high-quality (HQ) images from low-quality (LQ) ones with unknown degradations.
This problem has attracted widespread attention and achieved significant progress due to the development of deep learning methods.
However, it is still challenging to restore faithful SR images with high reality and fidelity, owing to the ill-posed property of SR problems.

Most state-of-the-art image SR methods rely on deep generative models, which aim to learn the distribution of HQ images and recover vivid textures from LQ inputs.
Generative Adversarial Networks (GANs), as one of the representative approaches, can recover sharp images with rich details~\cite{Real-ESRGAN, BSRGAN, GFPGAN, FeMaSR, GLEAN}.
However, the training process of GAN-based methods is unstable and the generated images usually exhibit perceptually unpleasant artifacts as pointed out by~\cite{FeMaSR, LDL, Desra}.
To alleviate these issues, several methods~\cite{FeMaSR, Codeformer} propose to construct a discrete codebook consisting of a pre-defined number of HQ feature vectors and learn to match LQ features to a set of HQ ones, which have achieved better perceptual quality.
Nevertheless, due to the vastness of the natural image space, priors encoded with a fixed size of codebook inherently possess a limited ability of representation, hindering faithful image SR results.

\begin{figure*}[!t]
\footnotesize
\centering
    \begin{tabular}{c c c c c c c}
            \multicolumn{3}{c}{\multirow{5}*[55.3pt]{
            \hspace{-2.5mm} \includegraphics[width=0.325\linewidth,height=0.275\linewidth]{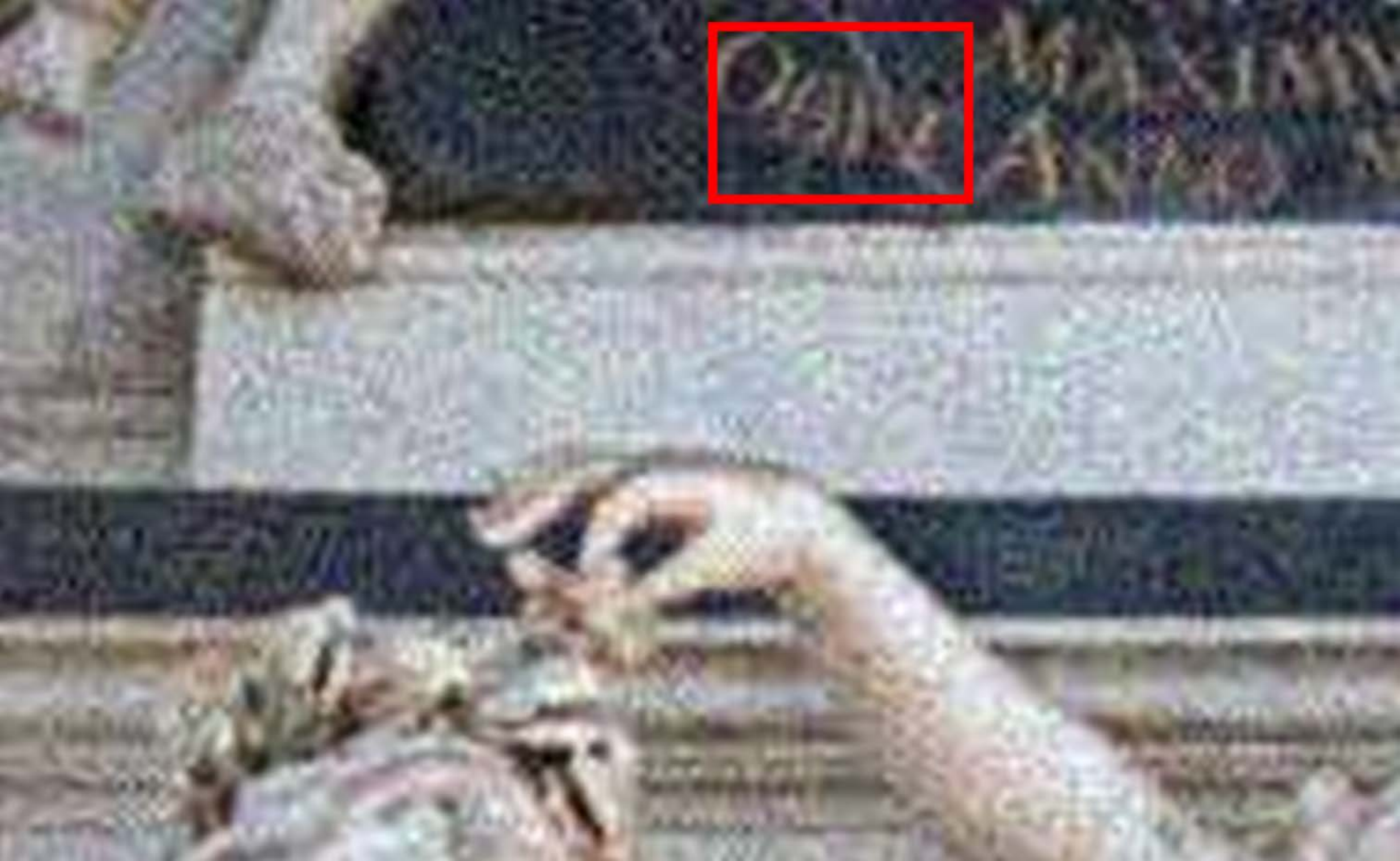}}}
            & \hspace{-4.0mm} \includegraphics[width=0.16\linewidth,height=0.125\linewidth]{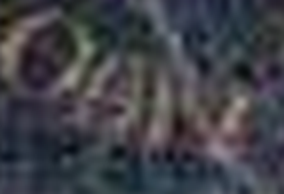}
            & \hspace{-4.0mm} \includegraphics[width=0.16\linewidth,height=0.125\linewidth]{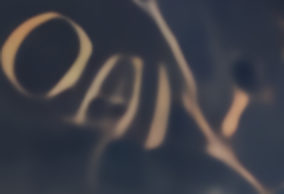}
            & \hspace{-4.0mm} \includegraphics[width=0.16\linewidth,height=0.125\linewidth]{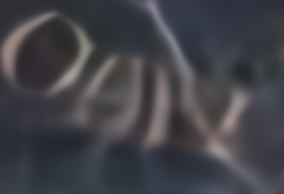}
            & \hspace{-4.0mm} \includegraphics[width=0.16\linewidth,height=0.125\linewidth]{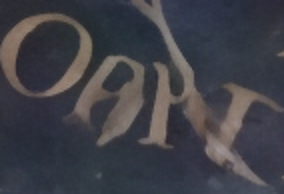}
              \\
    		\multicolumn{3}{c}{~}
            & \hspace{-4.0mm} (a) LQ patch
            & \hspace{-4.0mm} (b) DRM of DiffBIR
            & \hspace{-4.0mm} (c) DRM of SUPIR
            & \hspace{-4.0mm} (d) SeeSR~\cite{SeeSR} \\		
    	\multicolumn{3}{c}{~}
            & \hspace{-4.0mm} \includegraphics[width=0.16\linewidth,height=0.125\linewidth]{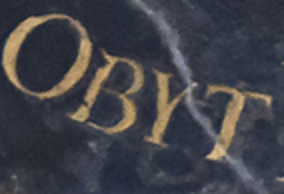}
            & \hspace{-4.0mm} \includegraphics[width=0.16\linewidth,height=0.125\linewidth]{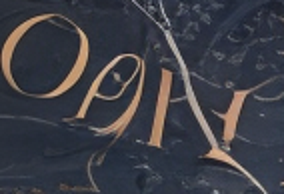}
            & \hspace{-4.0mm} \includegraphics[width=0.16\linewidth,height=0.125\linewidth]{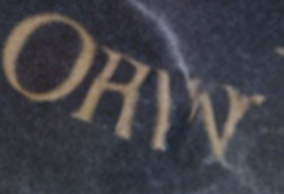}
            & \hspace{-4.0mm} \includegraphics[width=0.16\linewidth,height=0.125\linewidth]{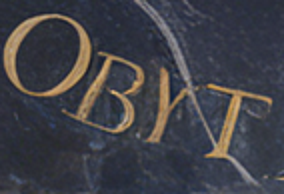}
            \\
    	\multicolumn{3}{c}{\hspace{-4.0mm} LQ image}
            & \hspace{-4.0mm} (e) GT patch
            & \hspace{-4.0mm} (f) DiffBIR~\cite{DiffBIR}
            & \hspace{-4.0mm} (g) SUPIR~\cite{SUPIR}
            & \hspace{-4.0mm} (h) Ours\\
    \end{tabular}
\vspace{-3mm}
    \caption{Visual comparison with state-of-the-art SR methods. (b) and (c) are the intermediate results of the degradation removal module (DRM) by DiffBIR~\cite{DiffBIR} and SUPIR~\cite{SUPIR}.
    The competing methods \cite{DiffBIR, SUPIR} do not effectively restore faithful structures based on the degradation removal results in (b)-(c).
    In contrast to the results in (d) and (f)-(g), our approach can recover more realistic high-quality results with faithful contents (\eg, the characters in (h)).
    }
\vspace{-3mm}
\label{fig:Intro_vis}
\end{figure*}

Recently, latent diffusion models (LDMs)~\cite{SD, SDXL, imagen, pixart} have achieved promising image generation results as they can effectively model complex image distributions.
While LDMs possess strong priors on the structures and details of HQ images, they face challenges in the image SR task, which requires not only vivid generation details but also faithful structure recovery that is consistent with LQ inputs.
To address this, several methods~\cite{DiffBIR, SUPIR, PASD, XPSR} enhance the encoder to extract degradation-robust features from LQ images and then leverage pre-trained diffusion models to refine the extracted features.
However, real-world images with unknown degradations present challenges for the encoder in extracting features with faithful structural information.
As the diffusion models are pre-trained on HQ images, any mistakes in the extracted features (Figure \ref{fig:Intro_vis}(b)-(c)) may be misinterpreted by the diffusion model as image structures, thus misleading the diffusion process and negatively impacting the final restoration results (\eg, the characters) in Figure \ref{fig:Intro_vis}(d) and (f)-(g).
As a result, relying solely on improving the extracted features for guiding the diffusion process has limitations in achieving faithful restoration.
Different from the above methods, we unleash and fine-tune the diffusion model to identify useful information from degraded inputs and boost faithful image SR.

Note that the encoder and the diffusion model play different but indispensable roles in LDM-based SR methods~\cite{SUPIR, DiffBIR}.
Separately optimizing these two modules will limit the power of LDMs, hindering the restoration of fine-scale structures as analyzed in Section \ref{sec: ablation_study}.
Thus, we propose to jointly optimize the encoder and diffusion model, establishing a more powerful LDM for image SR.
In addition, we develop a simple alignment module to help the encoder align well with the progressive diffusion process.
In this way, the encoder is able to provide useful latent features that coincide with the diffusion process and the diffusion model can further restore the features for better reconstruction.
Benefited from their interplay, the proposed method is able to distinguish between degradation effects and inherent structural information, and recover high-quality and faithful results, as shown in Figure~\ref{fig:Intro_vis}(h).

The contributions of this work are summarized as follows:
\emph{(i)} We propose an effective method, named FaithDiff, which unleashes diffusion priors to better harness the powerful representation ability of LDM for image SR.
A detailed analysis demonstrates that unleashing diffusion priors is more effective in exploring useful information from degraded inputs, suppressing reconstruction errors, and recovering faithful structures.
\emph{(ii)} We develop a simple yet effective alignment module to align the latent representation of LQ inputs, encoded by the encoder, with the noisy latent of the diffusion model.
\emph{(iii)} We jointly fine-tune the encoder and the diffusion model to benefit from their interplay.
\emph{(iv)} We quantitatively and qualitatively demonstrate that our {FaithDiff} performs favorably against state-of-the-art methods on both synthetic and real-world benchmarks.

\section{Related Work}
\textbf{Image super-resolution.} Image SR is a highly ill-posed problem, and early approaches~\cite{SRMD, unfoldingSR, IKC} focus primarily on developing methods to estimate degradation kernels for restoration.
However, due to the complexity of degradation in real-world scenarios, these methods often fail to remove degradation artifacts and recover sharp edges effectively.
To better solve real-world degradation effects, Zhang et al.~\cite{Real-ESRGAN} and Wang et al.~\cite{BSRGAN} consider more complex degradation scenarios and propose high-order degradation models to address real-world degradations.

\begin{figure*}[t]
	\begin{center}
		\includegraphics[width=0.98\linewidth]{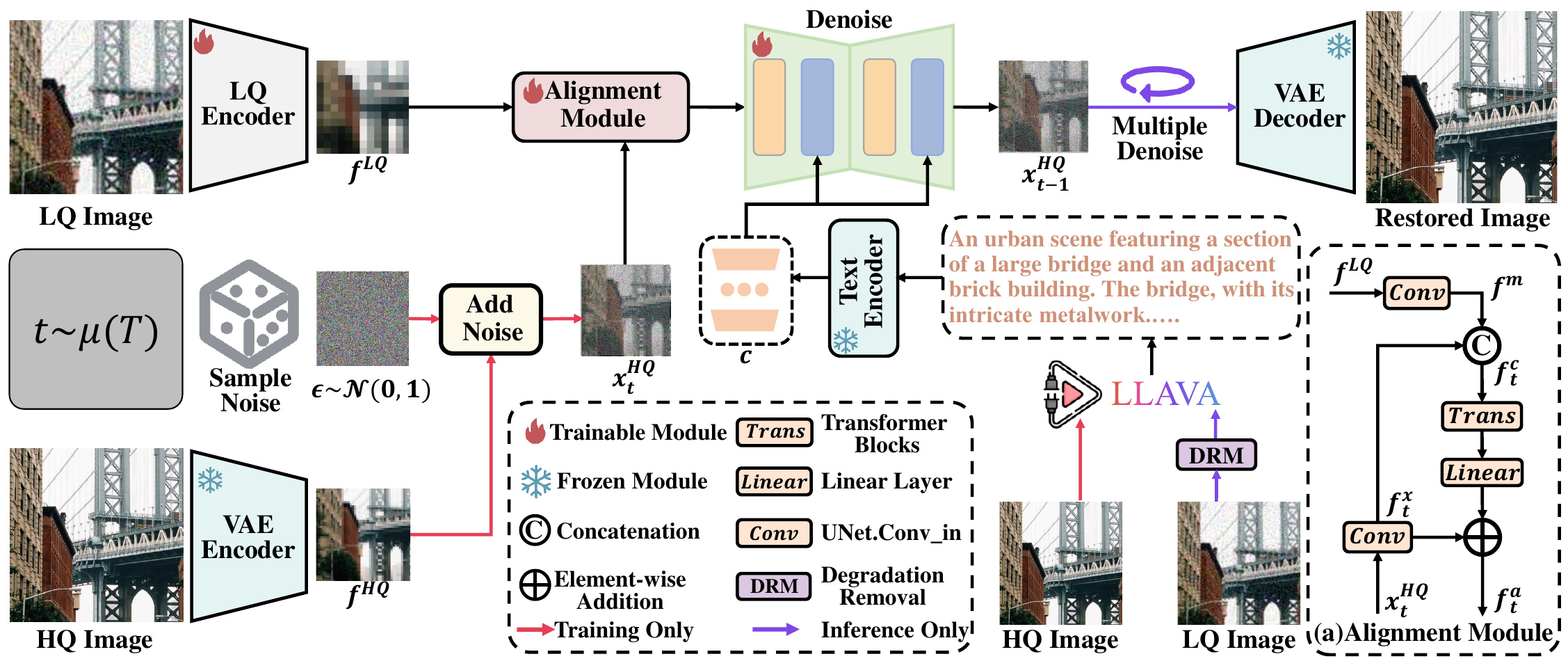}
	\end{center}
        \vspace{-2mm}
	\caption{\textbf{An overview of the FaithDiff}, which takes LQ images and image descriptions as inputs and restores HQ images via diffusion process.
    To fully leverage the power of LDMs, we propose to unleash diffusion priors.
    An alignment module is developed to effectively incorporate the features extracted from LQ images with the noisy latent of the diffusion model.
    We jointly optimize the encoder, the alignment module, and the diffusion model, which can benefit from their interplay and lead to faithful SR images with high viusal quality.
    }
	\label{fig: method_pipeline}
	\vspace{-2mm}
\end{figure*}

By incorporating high-order degradation models, several GAN-based methods~\cite{Real-ESRGAN, BSRGAN, FeMaSR, DASR} provide promising visual results.
However, GAN-based approaches often suffer from perceptually unpleasant artifacts. Liang et al.~\cite{LDL} and Xie et al.~\cite{Desra} attempt to mitigate this issue by penalizing regions with such artifacts.
Nonetheless, these
methods still struggle to recover fine details in LQ images.

With the success of generative large models such as Stable Diffusion~\cite{SD, SDXL}, several recent methods leverage generative priors to solve image SR problem~\cite{StableSR, DiffBIR, PASD, SUPIR, XPSR, SeeSR}.
Yang et al.~\cite{PASD} introduce a degradation removal module that extracts degradation-insensitive features from LQ images to guide the diffusion process for
restoration.
Lin et al.~\cite{DiffBIR} explicitly divide the restoration process into two stages: degradation removal and detail regeneration.
It first uses MSE-based restoration models to achieve degradation removal and then employs generative priors for detail enhancement.
Yu et al.~\cite{SUPIR} also adopt a similar two-stage restoration approach.
Although promising results have been achieved, the performance of these algorithms is limited by the accuracy of degradation removal results.

\noindent \textbf{Diffusion model.}
Recently, Denoising Diffusion Probabilistic Models (DDPM)~\cite{DDPM} demonstrate remarkable capabilities in generating high-quality natural images.
Rombach et al.~\cite{SD} extend the DDPM framework to the latent space, achieving impressive results in text-to-image synthesis.
This advancement leads to the proliferation of large pre-trained text-to-image diffusion models, such as Stable Diffusion (SD)~\cite{SD}, Imagen~\cite{imagen}, and PixelArt-$\alpha$~\cite{pixart}.
Several studies~\cite{HLDM, HCL} show that text-to-image diffusion priors are more effective than GAN priors in handling diverse natural images.
Lv et al.~\cite{controlnet} and Mou et al.\cite{T2I_adapter} use edge maps, segmentation maps, etc., as additional inputs to better control the diffusion process to generate specified content.
Hu et al.~\cite{Animate} and Tian et al.~\cite{EMO} achieve promissing results in human pose deformation and facial animation by leveraging pre-trained diffusion models.
While these methods generate visually diverse output, they face challenges in image SR task, which require recovering faithful structural details from LQ images.

\section{FaithDiff}
Our goal is to present an effective method to fully harness the power of LDM for image SR.
Specifically, we first employ the encoder of a pre-trained Variational Autoencoder~\cite{Autoencoder} (VAE) to map LQ inputs into the latent space and extract the corresponding LQ features.
Then, we develop an alignment module to effectively transfer useful information from the latent LQ features and ensure them align well with the diffusion process.
In addition, we incorporate text embeddings, extracted from image descriptions using a pre-trained text encoder, as auxiliary information. These embeddings are integrated with the latent features of the diffusion model through cross-attention layers to help explore useful structural information.
Furthermore, we propose a unified feature optimization strategy to jointly fine-tune the VAE encoder and the diffusion model, allowing the encoder to extract useful information from LQ images that facilitates the diffusion process while enabling the diffusion model to further refine the extracted feature for HQ image SR.
Finally, we obtain the restored image from the refined features by a pre-trained VAE decoder.
Figure~\ref{fig: method_pipeline} summarizes the network architecture.
In the following, we present the details of the proposed approach.

\subsection{LQ feature extraction}
Given an LQ image, we extract the LQ features $f^{LQ}$ through a VAE encoder.
Existing methods~\cite{SUPIR,DiffBIR} mostly adopt the features extracted from the last layer of the VAE encoder as the LQ features.
However, the last layer of the encoder significantly compresses the channel dimensions of the features (\eg, only $8$ channels used in~\cite{SUPIR,DiffBIR}), which is insufficient to capture both the degradation factors and structural details in LQ inputs.
To obtain sufficient feature information, we propose to employ the features from the penultimate layer of the encoder as $f^{LQ}$, which have $512$ channels and contain more useful information that is beneficial for the subsequent diffusion process (as demonstrated in Section \ref{sec: ablation_study}).

\subsection{Alignment module}
To effectively transmit the information contained in the extracted LQ features $f^{LQ}$ to control the generation of the diffusion model, a straightforward solution is to add $f^{LQ}$ with the noisy latent $x_t^{HQ}$ of the diffusion model after passing them through convolutional layers, respectively.
However, as the diffusion process progresses, the $x_t^{HQ}$ become increasingly clear, while the $f^{LQ}$ remain unchanged.
Therefore, it is unreasonable to directly combine them, as the degradation factors in $f^{LQ}$ may continuously interfere with the generation of high-quality  $x_t^{HQ}$.

To this end, we propose an effective alignment module that extracts useful features from $f^{LQ}$ to achieve better alignment with $x_t^{HQ}$, thus facilitating the subsequent diffusion process.
The architecture of the proposed alignment module is illustrated in Figure~\ref{fig: method_pipeline} (a).

First, we individually employ convolutional operations on the LQ features $f^{LQ}$ as well as the noisy latent $x_t^{HQ}$, and obtain their concatenation result as $f^c_t$.
Then, $f^c_t$ is put into a stack of two Transformer blocks~\cite{Transformer}, which facilitates the interaction between the LQ features $f^{LQ}$ and the noisy latent $x_t^{HQ}$.
Finally, we obtain the aligned features $f^a_t$ as
\begin{equation}
    \begin{split}
    & f^x_t = \text{Conv}(x_t^{HQ}), f^{m}=\text{Conv}(f^{LQ}),\\
    & f^c_t = \text{Concat}(f^x_t, f^{m}),\\
    & \text{Trans}(f^c_t) = \mathcal{T}_2(\mathcal{T}_1(f^c_t)), \\
    & f^{a}_t = \text{Linear}(\text{Trans}(f^c_t) + f^x_t),
    \end{split}
\end{equation}
where $\mathcal{T}_i~ (i=1,2)$ denotes the transformer block~\cite{Transformer}, $\text{Conv}(\cdot)$ denotes the $3 \times 3$ convolution, $\text{Linear}(\cdot)$ denotes the fully connected layer, and
$\text{Concat}(\cdot)$ denotes the concatenation operation.

%
Note that our alignment module is simple yet effective.
More advanced architectures can also be considered to align the LQ features with the noisy latent of diffusion model.

During the diffusion process, we also leverage text embeddings as additional auxiliary information to help extract useful information from latent features of the diffusion model.
These text embeddings are extracted from image descriptions by a pre-trained text encoder~\cite{CLIP}.
Similar to \cite{SUPIR,PASD}, we utilize the cross-attention layer to establish interactions between text embeddings and latent features, capturing the text-to-image knowledge inherent in LDM.
Overall, given the latent HQ features $x_t^{HQ}$ and the aligned features $f^{a}_t$ at the diffusion step $t$ as well as the text embeddings $c$, we can obtain the latent HQ features at the diffusion step $t-1$ as
\begin{equation}
    \begin{split}
        x_{t-1}^{HQ} = \frac{1}{\sqrt{\alpha_t}}(x_t^{HQ} - \frac{1-\alpha_t}{\sqrt{1-\bar \alpha_t}} \hat{\epsilon}_\theta (f^{a}_t, c, t)) + \sigma_t z,
    \end{split}
    \label{eq: alignment}
\end{equation}
where $\hat{\epsilon}_\theta$ denotes the diffusion model, $z$ is sampled from the standard normal distribution $\mathcal{N}(0,1)$, $\alpha_t$ and $\sigma_t$ are the noisy scheduler terms that control sample quality, and $\bar{\alpha}_t = \prod_{i=1}^t \alpha_i$.

\subsection{Unified feature optimization}
Our FaithDiff contains an LQ feature extraction module (\ie, a VAE encoder), a diffusion module, an alignment module to incorporate the extracted LQ features with the noisy latent of the diffusion model, a text embedding extraction module (\ie, a text encoder~\cite{CLIP}) and a HQ image reconstruction module (\ie, a VAE decoder).
We unify the VAE encoder, alignment module, and diffusion model into a trainable framework while keeping the other components frozen due to the VAE decoder's strong priors for reconstructing HQ images and the text encoder's robust text-image alignment capabilities.
To connect LQ features extracted from VAE encoder with noisy latent, we first only pre-train the alignment module, while fixing the VAE encoder and the diffusion model.
Then we jointly fine-tune the VAE encoder, the alignment module, and the diffusion model, in order to facilitate these three modules adapting to each other and leveraging their interplays.
The fine-tuned VAE encoder is denoted as LQ encoder in Figure~\ref{fig: method_pipeline}.

The proposed method is optimized by:
\begin{equation}
L = \left\| \epsilon - \hat{\epsilon}_\theta (\sqrt{\bar{\alpha}_t x^{HQ}_0} + \sqrt{1 - \bar{\alpha}_t} \epsilon, f^{LQ}, c, t) \right\|_1,
\end{equation}
where $\left\|\cdot\right\|_1$ is the L1 normalization, $\epsilon$ is sampled from the standard normal distribution $\mathcal{N}(0, \mathbf{I})$, $x^{HQ}_0$ is generated from the HQ image by the pre-trained VAE encoder, $\hat{\epsilon}_\theta$, $\bar{\alpha}_t$, $c$, and $t$ are defined the same as those in \eqref{eq: alignment}.

\begin{table*}[!t]
\centering
\renewcommand{\baselinestretch}{1.2}
\caption{\setstretch{0.9} Quantitative comparison with state-of-the-art methods on the datasets of DIV2K-Val~\cite{DIV2K} and LSDIR-Val~\cite{LSDIR}. `Level I', `Level II', and `Level III' denote the datasets with mild, medium, and severe degradations, respectively. The best and second performances are marked in {\color{red}red} and {\color{blue}blue}, respectively.}
\vspace{-2mm}

\resizebox{\textwidth}{!}{%
\begin{tabular}{c|l|ccccc||ccccc}
\toprule
        \multirow{2}{*}{D-Level} & \multirow{2}{*}{Methods}     & \multicolumn{5}{c||}{DIV2K-Val~\cite{DIV2K}}         & \multicolumn{5}{c}{LSDIR-Val~\cite{LSDIR}}       \\ \cline{3-12}
 &  & PSNR (dB)~{\color{red}$\uparrow$} & SSIM~{\color{red}$\uparrow$} & LPIPS~{\color{red}$\downarrow$} & MUSIQ~{\color{red}$\uparrow$} & CLIPIQA+~{\color{red}$\uparrow$} & PSNR (dB)~{\color{red}$\uparrow$} & SSIM~{\color{red}$\uparrow$} & LPIPS~{\color{red}$\downarrow$} & MUSIQ~{\color{red}$\uparrow$} & CLIPIQA+~{\color{red}$\uparrow$} \\ \hline
           & Real-ESRGAN~\cite{Real-ESRGAN}           & {\color{blue}26.64} & {\color{blue}0.7737} & {\color{red}0.1964}         & 62.38        & 0.4649 & {\color{blue}23.47} & {\color{blue}0.7102} & {\color{red}0.2008}         & 69.23        & 0.5378 \\
 & BSRGAN~\cite{BSRGAN}                & {\color{red}27.63} & {\color{red}0.7897} & {\color{blue}0.2038}         & 61.81        & 0.4588 & {\color{red}24.42} & {\color{red}0.7292} & {\color{blue}0.2167}         & 66.21        & 0.5037 \\
 & StableSR~\cite{StableSR}              &  24.71     & 0.7131       & 0.2393                &  65.55            & 0.5156       & 21.57 & 0.6233 & 0.2509         & 70.52        & 0.6004 \\
Level-I   & DiffBIR~\cite{DiffBIR}                      & 24.60 & 0.6595 & 0.2496 & 66.23 & 0.5407   & 21.75 & 0.5837 & 0.2677 & 68.96 & 0.5693   \\
 & PASD~\cite{PASD}                  & 25.31 & 0.6995 & 0.2370         & 64.57        & 0.4764 & 22.16 & 0.6105 & 0.2582         & 68.90        & 0.5221 \\
  & SeeSR~\cite{SeeSR}                  & 25.08      & 0.6967       & 0.2263                &  {\color{blue}66.48}           & 0.5336       & 22.68      & 0.6423       & 0.2262               & 70.94             & 0.5815       \\
 & DreamClear~\cite{dreamclear}                  & 23.76      & 0.6574       & 0.2259                &  66.15            & {\color{red}0.5478}       & 20.08      & 0.5493       & 0.2619               & 70.81             & {\color{blue}0.6182}       \\
 & SUPIR~\cite{SUPIR}                 & 25.09 & 0.7010 & 0.2139         & 65.49        & 0.5202 & 21.58 & 0.5961 & 0.2521         & {\color{blue}71.10}        & 0.6118 \\
 & Ours                  & 24.29      &  0.6668      &   0.2187             & {\color{red}66.53}        & {\color{blue}0.5432} & 21.20 & 0.5760 & 0.2264         &  {\color{red}71.25}            & {\color{red}0.6253}       \\ \hline
 & Real-ESRGAN~\cite{Real-ESRGAN}           & {\color{blue}25.49} & {\color{blue}0.7274} & {\color{red}0.2309}         & 61.84        & 0.4719 & {\color{blue}22.47} & {\color{blue}0.6567} & {\color{red}0.2342}         & 69.14        & 0.5456 \\
 & BSRGAN~\cite{BSRGAN}                & {\color{red}26.42} & {\color{red}0.7402} & 0.2465         & 60.00        & 0.4463 & {\color{red}23.35} & {\color{red}0.6682} & 0.2641         & 64.17        & 0.4858 \\
 & StableSR~\cite{StableSR}              & 24.26 & 0.6771 & 0.2590         & 64.76        & 0.5057 & 21.58 & 0.5946 & 0.2802         & 69.57        & 0.5667 \\
Level-II  & DiffBIR~\cite{DiffBIR}                      & 24.42 & 0.6441 & 0.2708  & 64.83 & 0.5246   & 21.63 & 0.5672 & 0.2853 & 67.61 & 0.5555   \\
 & PASD~\cite{PASD}                  & 24.89 & 0.6764 & 0.2502         & 64.45        & 0.4718 & 21.85 & 0.5846 & 0.2737         & 68.53        & 0.5131 \\
 & SeeSR~\cite{SeeSR}                  & 24.65      & 0.6734       & 0.2428                &  {\color{blue}66.09}           & 0.5226       & 22.00      & 0.6026       & {\color{blue}0.2469}               & {\color{blue}70.91}             & 0.5837       \\
 & DreamClear~\cite{dreamclear}                  & 23.39      & 0.6330       & 0.2518               &  64.96            &  {\color{blue}0.5295}      & 19.74      & 0.5191       & 0.2910               & 70.41             & {\color{blue}0.6072}        \\
 & SUPIR~\cite{SUPIR}                 & 24.42 & 0.6703 & 0.2432         & 65.58        & 0.5202 & 21.30 & 0.5713 & 0.2733         & 70.59        & 0.5998 \\
 & Ours                  & 23.80 & 0.6413 & {\color{blue}0.2407}         & {\color{red}66.42}        & {\color{red}0.5460} & 20.88 & 0.5493 & {\color{blue}0.2469}         &  {\color{red}71.15}    & {\color{red}0.6219}       \\ \hline
 & Real-ESRGAN~\cite{Real-ESRGAN}           & 22.81 & {\color{red}0.6288} & 0.3535         & 60.11        & 0.4637 & 20.13 & {\color{blue}0.5374} & 0.3650         & 67.02        & 0.5275 \\
 & BSRGAN~\cite{BSRGAN}                & {\color{red}23.45} & {\color{blue}0.6281} & 0.3462         & 62.41        & 0.4838 & {\color{red}20.75} & {\color{red}0.5358} & 0.3667         & 67.41        & 0.5363 \\
 & StableSR~\cite{StableSR}              & 23.34 & 0.6277 & 0.3559         & 57.89        & 0.4124 & {\color{blue}20.55} & 0.5195 & 0.3716         & 64.31        & 0.4859 \\
Level-III & DiffBIR~\cite{DiffBIR}                      & {\color{blue}23.42} & 0.5992 & 0.3676 & 58.86 & {\color{blue}0.5154}   & 20.53 & 0.4809 & 0.3951 & 62.23 & 0.5148   \\
 & PASD~\cite{PASD}                  & 22.58 & 0.5985 & 0.3646         & 63.08        & 0.4815 & 20.03 & 0.4974 & 0.3769         & 67.43        & 0.5092 \\
 & SeeSR~\cite{SeeSR}                  & 22.58      & 0.5944       & 0.3278                &  {\color{blue}65.82}           & 0.5106       & 20.16      & 0.5046       & {\color{blue}0.3437}               & 69.35             & 0.5444    \\
 & DreamClear~\cite{dreamclear}                  & 21.82      & 0.5510       & 0.3336               & 62.59             & 0.4914       & 18.46     & 0.4341       &  0.3831              &  68.64            &  0.5757      \\
 & SUPIR~\cite{SUPIR}                 & 21.90 & 0.5611 & {\color{blue}0.3172}         & 65.46        & 0.5134 & 19.17 & 0.4650 & 0.3488         & {\color{blue}70.16}        & {\color{blue}0.5917} \\
 & Ours                  & 21.77 & 0.5662 & {\color{red}0.3080}         & {\color{red}66.28}        & {\color{red}0.5275} & 18.92 & 0.4568 & {\color{red}0.3170}         & {\color{red}71.37}        & {\color{red}0.6067}  \\ \bottomrule
\end{tabular}}
\label{tab: synthetic_comp}
\vspace{-2mm}
\end{table*}

\section{Experimental Results}
In this section, we first describe the datasets and implementation details of the proposed method.
Then we evaluate our approach against state-of-the-art ones using publicly available benchmark datasets.
More experimental results are included in the supplemental material.

\subsection{Datasets and implementation details}
\label{Sec: datasets}
\noindent \textbf{Training dataset.}
We collect a training dataset of high-resolution images from LSDIR~\cite{LSDIR}, DIV2K~\cite{DIV2K}, Flicker2K \cite{Flicker2k}, DIV8K~\cite{div8k}, and $10,000$ face images from FFHQ~\cite{FFHQ}.
We generate LQ images following the same configuration as~\cite{PASD}.
Similar to \cite{SUPIR}, we use the LLAVA~\cite{LLAVA} to generate textual descriptions for each image.

\noindent \textbf{Synthetic test datasets.}
\label{synthesis}
Similar to DASR~\cite{DASR}, we employ different levels of degradations (D-level) to synthesize degraded validation sets of DIV2K~\cite{DIV2K} and LSDIR~\cite{LSDIR}.

\noindent \textbf{Real-world test datasets.}
To evaluate our approach in real scenarios, we first test on the dataset of RealPhoto60~\cite{SUPIR}.
We then collect a dataset, named RealDeg, of $238$ images including old photographs, classic film stills, and social media images to evaluate our method across diverse degradation types.
More details are included in the supplemental material.

\noindent \textbf{Implementation details.}
We choose the base model of SDXL~\cite{SDXL} as our diffusion model and use its VAE encoder as our LQ encoder.
We train the proposed method using two A800 GPUs and adopt the AdamW optimizer~\cite{Adamw} with default parameters.
We crop images into patches of $512 \times 512$ pixels during training and set the batch size as $256$.
We first pre-train the proposed alignment module for $6,000$ iterations with an initial learning rate of $5\times10^{-5}$.
We then fine-tune the whole network including the LQ encoder, the alignment module, and the diffusion model in an end-to-end manner for $40,000$ iterations.
During the fine-tuning stage, the learning rate for the LQ encoder and other network components are initialized as $5\times10^{-6}$ and $10^{-5}$, respectively. The learning rates are updated using the Cosine Annealing scheme~\cite{cosine}.
To further enhance the controllability of our proposed method, we employ an image descriptions dropout operation in training to enable classifier-free guidance (CFG)~\cite{classifier}.
We empirically set the image descriptions dropout ratio to 20\% during training.
For inference of FaithDiff, we adopt Euler scheduler~\cite{elucidating} with $20$ sampling steps and set CFG guidance scale as $5$.

\subsection{Comparisons with the state of the art}
We compare our approach with state-of-the-art image SR methods including GAN-based methods (\ie, Real-ESRGAN~\cite{Real-ESRGAN} and BSRGAN~\cite{BSRGAN}) and diffusion-based methods (\ie, StableSR~\cite{StableSR}, DiffBIR~\cite{DiffBIR}, PASD~\cite{PASD}, DreamClear~\cite{dreamclear}, and SUPIR~\cite{SUPIR}).
We use PSNR, SSIM and perceptual-oriented metrics (\ie, LPIPS~\cite{Lpips}, MUSIQ~\cite{MUSIQ}, and CLIPIQA+~\cite{CLIPIQA}) to evaluate the fidelity and quality of restored images.

\begin{figure*}[!t]
\footnotesize
\centering
    \begin{tabular}{c c c c c c c}
            \multicolumn{3}{c}{\multirow{5}*[45.6pt]{
            \hspace{-2.5mm} \includegraphics[width=0.325\linewidth,height=0.235\linewidth]{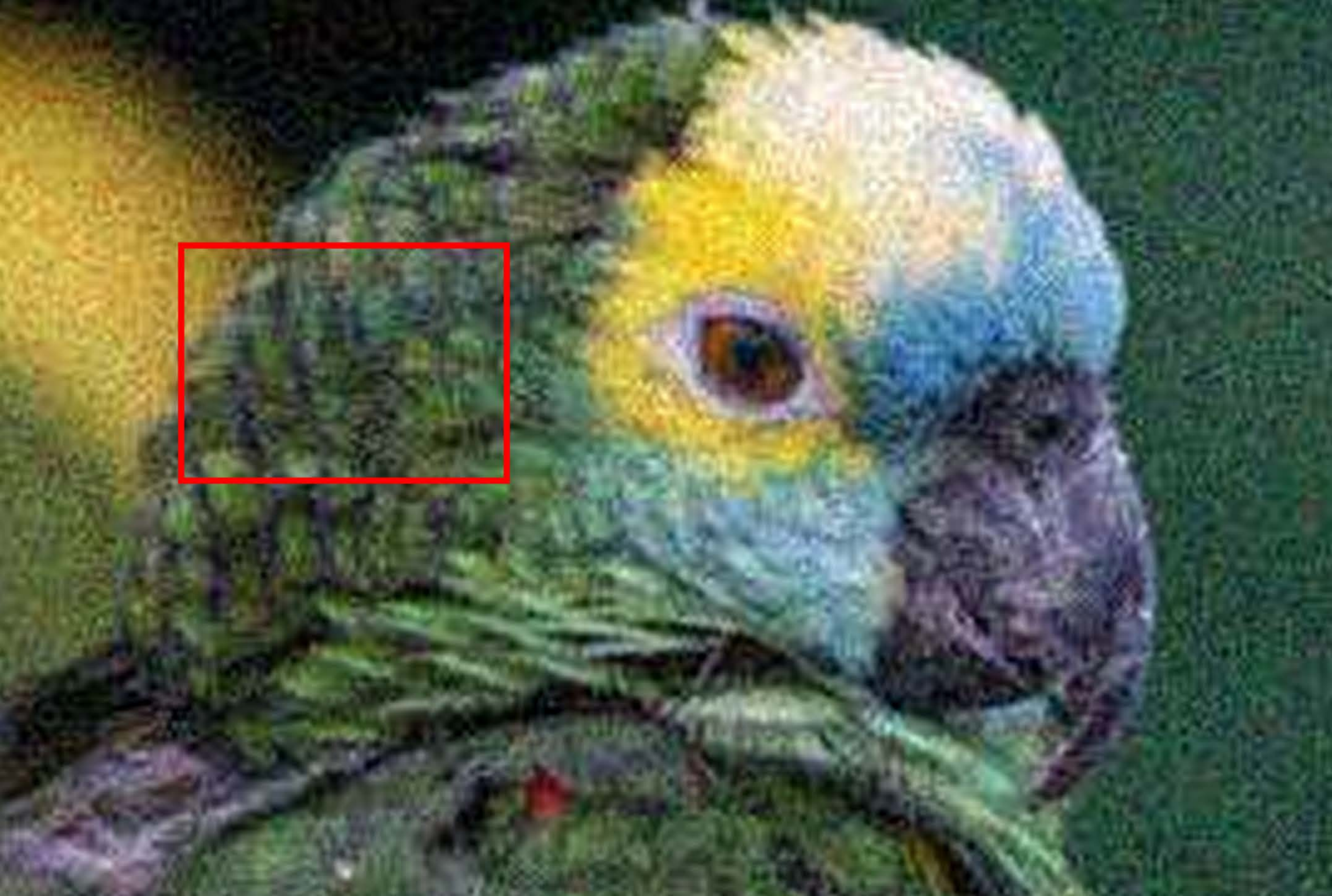}}}
            & \hspace{-4.0mm} \includegraphics[width=0.16\linewidth,height=0.105\linewidth]{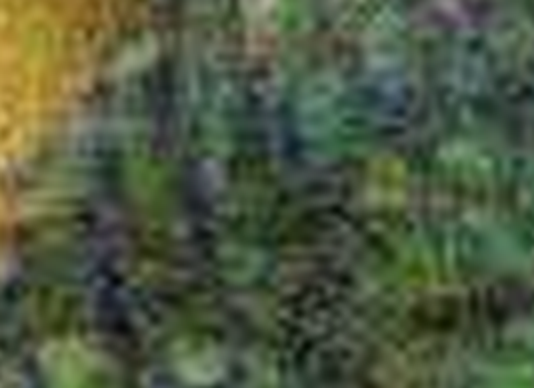}
            & \hspace{-4.0mm} \includegraphics[width=0.16\linewidth,height=0.105\linewidth]{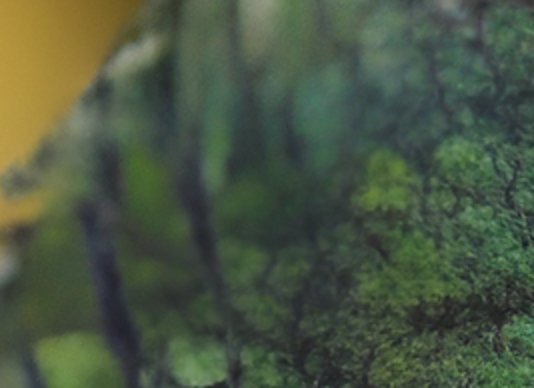}
            & \hspace{-4.0mm} \includegraphics[width=0.16\linewidth,height=0.105\linewidth]{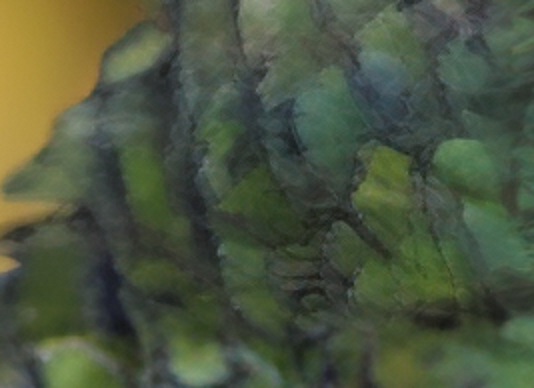}
            & \hspace{-4.0mm} \includegraphics[width=0.16\linewidth,height=0.105\linewidth]{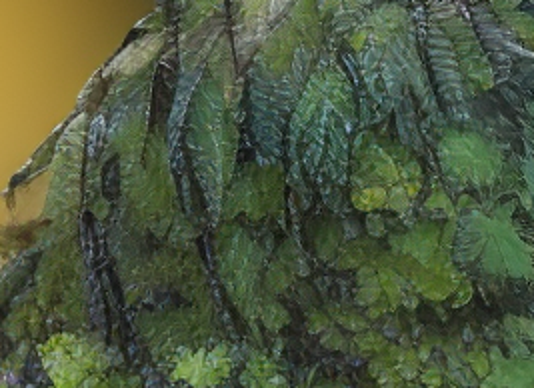}
              \\
    		\multicolumn{3}{c}{~}
            & \hspace{-4.0mm} (a) LQ patch
            & \hspace{-4.0mm} (b) Real-ESRGAN~\cite{Real-ESRGAN}
            & \hspace{-4.0mm} (c) StableSR~\cite{StableSR}
            & \hspace{-4.0mm} (d) DiffBIR~\cite{DiffBIR} \\		
    	\multicolumn{3}{c}{~}
            & \hspace{-4.0mm} \includegraphics[width=0.16\linewidth,height=0.105\linewidth]{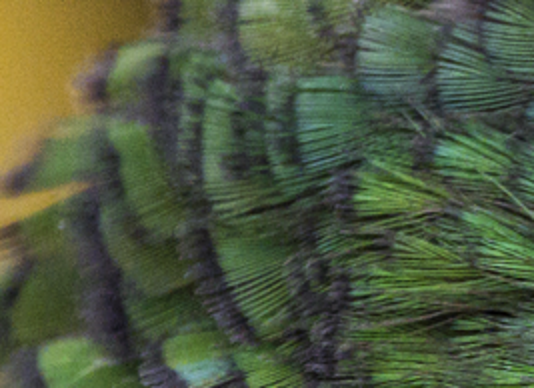}
            & \hspace{-4.0mm} \includegraphics[width=0.16\linewidth,height=0.105\linewidth]{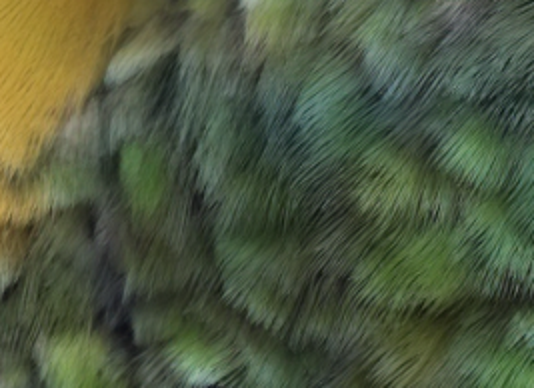}
            & \hspace{-4.0mm} \includegraphics[width=0.16\linewidth,height=0.105\linewidth]{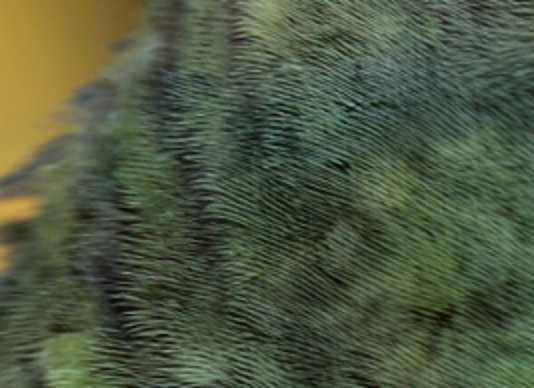}
            & \hspace{-4.0mm} \includegraphics[width=0.16\linewidth,height=0.105\linewidth]{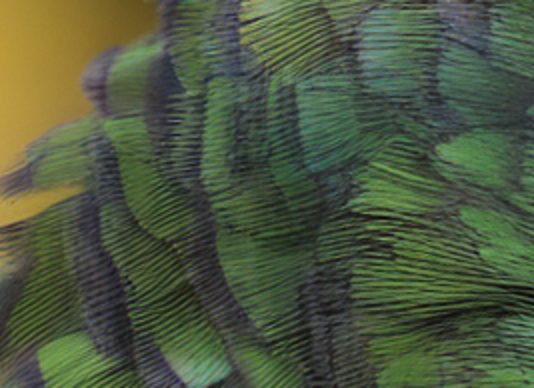}
            \\
    	\multicolumn{3}{c}{\hspace{-4.0mm} LQ image from LSDIR~\cite{LSDIR}}
            & \hspace{-4.0mm} (e) GT Patch
            & \hspace{-4.0mm} (f) SeeSR~\cite{SeeSR}
            & \hspace{-4.0mm} (g) SUPIR~\cite{SUPIR}
            & \hspace{-4.0mm} (h) Ours\\

    \end{tabular}
\vspace{-2mm}
\caption{Image SR result (×4) on the synthetic benchmark. The restored image by GAN-based methods~\cite{Real-ESRGAN} exhibits perceptually unpleasant artifacts in (b). Existing diffusion-based methods~\cite{StableSR,DiffBIR,SeeSR,SUPIR} over-smooth the details in (c) or generate incorrect structures in (d) and (f)-(g). In contrast, the proposed method recovers much clearer images with faithful structures in (h).}
\label{fig: vis_comp}
\vspace{-2mm}
\end{figure*}

\begin{table*}[!t]
\centering
\caption{Quantitative comparison with state-of-the-art methods on real-world benchmarks. The best and second performances are marked in {\color{red}red} and {\color{blue}blue}, respectively.}
\vspace{-2mm}
\resizebox{\textwidth}{!}{
\begin{tabular}{c|c|ccccccccc}
\toprule
Benchmarks                & Metrics  & Real-ESRGAN~\cite{Real-ESRGAN} & BSRGAN~\cite{BSRGAN} & StabeSR~\cite{StableSR} & DiffBIR~\cite{DiffBIR} & PASD~\cite{PASD}   & SeeSR~\cite{SeeSR} & DreamClear~\cite{dreamclear} & SUPIR~\cite{SUPIR}  & Ours   \\ \hline
\multirow{2}{*}{RealPhoto60~\cite{SUPIR}}   & MUSIQ~{\color{red}$\uparrow$}    &  59.29      & 45.46  & 57.89   & 63.67   & 64.53  & {\color{blue}70.80}  & 70.46      & 70.26  & {\color{red}72.74}  \\
                          & CLIPIQA+~{\color{red}$\uparrow$} & 0.4389      & 0.3397 & 0.4214  & 0.4935  & 0.4786 & {\color{blue}0.5691}  & 0.5273     & 0.5528 & {\color{red}0.5932} \\ \hline
\multirow{2}{*}{RealDeg}  & MUSIQ~{\color{red}$\uparrow$}    & 52.64       & 52.08  & 53.53   & 58.22   & 47.31  & {\color{blue}60.10}  & 56.67 & 51.50  & {\color{red}61.24}  \\
    & CLIPIQA+~{\color{red}$\uparrow$} & 0.3396      & 0.3520 & 0.3669  & 0.4258  & 0.3137 & {\color{blue}0.4315} & 0.4105 & 0.3468 & {\color{red}0.4327} \\ \bottomrule
\end{tabular}}
\label{tab:real_world_comp}
\vspace{-2mm}
\end{table*}
\begin{table*}[!t]
\centering
\caption{Recognition results on the dataset of Occluded RoadText 2024~\cite{ICDAR} (OCR recognition). The best and second performances are highlighted in {\color{red}red} and {\color{blue}blue}, respectively.}
\vspace{-2mm}
	   	   \resizebox{\textwidth   }{!}{

\begin{tabular}{c|cc|ccccccccc}
\toprule
Metrics   & GT & LQ & Real-ESRGAN~\cite{Real-ESRGAN} & BSRGAN~\cite{BSRGAN}  & StableSR~\cite{StableSR} & DiffBIR~\cite{DiffBIR} & PASD~\cite{PASD} & SeeSR~\cite{SeeSR} & DreamClear~\cite{dreamclear} & SUPIR~\cite{SUPIR} & Ours \\ \hline
Precision & 52.72\%   & 7.54\%   & 13.19\%                    &  12.04\%      &  19.87\%        & 26.21\%        & 24.32\% & 30.07\%   & 22.45\%     & {\color{blue}31.78\%}      & {\color{red}36.45\%}     \\
Recall    & 56.67\%   & 7.59\%   & 13.68\%                   & 12.33\%       &  20.31\%        &  27.90\%       & 25.14\%  & 33.09\%  & 23.50\%     & {\color{blue}41.57\%}      &  {\color{red}46.74\%}    \\ \bottomrule
\end{tabular}}
\vspace{-2mm}
\label{tab: ocr}
\end{table*}

\noindent \textbf{Evaluations on the synthetic datasets.}
We first evaluate our approach on the datasets of DIV2K~\cite{DIV2K} and LSDIR~\cite{LSDIR}.
Table~\ref{tab: synthetic_comp} shows the quantitative results.
Although GAN-based methods~\cite{Real-ESRGAN, BSRGAN} perform best in terms of PSNR and SSIM metrics, they exhibit poor performance in terms of MUSIQ~\cite{MUSIQ} and CLIPIQA+~\cite{CLIPIQA} metrics.
The proposed approach outperforms competing methods~\cite{Real-ESRGAN, BSRGAN, StableSR, DiffBIR, PASD, SeeSR, dreamclear, SUPIR} in terms of the perceptual-oriented metrics.
Note that our method clearly improves the results, especially in the challenging cases of severe degradations, which demonstrates the effectiveness of our approach.

Figure~\ref{fig: vis_comp} shows visual comparisons on an LQ images with severe degradations from LSDIR~\cite{LSDIR}.
The results generated by GAN-based methods~\cite{Real-ESRGAN} exhibit perceptually unpleasant artifacts as shown in Figure~\ref{fig: vis_comp}(b).
Existing diffusion-based methods~\cite{StableSR,DiffBIR,SeeSR,SUPIR}, however, also do not recover satisfying results.
The method~\cite{StableSR} that directly injects LQ input into the diffusion process generates over-smoothed regions, as shown in Figure~\ref{fig: vis_comp}(c), as frozen generative priors are difficult to align well with LQ input, which limits the powerful generative capability.
Although the methods~\cite{SeeSR, DiffBIR, SUPIR} propose to extract degradation-robust features from LQ inputs, any mistake in the LQ features can mislead the diffusion process to generate inaccurate results (\eg, severe artifacts obtained in Figure~\ref{fig: vis_comp}(f) and incorrect structures generated in Figure~\ref{fig: vis_comp}(d) and (g)).
In contrast, by unleashing the diffusion prior and jointly optimizing it with the encoder, our approach yields more realistic image with faithful structural details (\eg, the restored feathers of a bird in Figure~\ref{fig: vis_comp}(h)) that are consistent with the ground truth as shown in Figure~\ref{fig: vis_comp}(e).

\noindent \textbf{Evaluations on the real-world datasets.}
We further evaluate the proposed method on real-world scenarios. As high-resolution images are not available, we use non-reference metrics (\ie, MUSIQ~\cite{MUSIQ} and CLIPIQA+~\cite{CLIPIQA}) to evaluate the quality of restored images in Table~\ref{tab:real_world_comp}, where the proposed approach achieves the best performance, increasing MUSIQ~\cite{MUSIQ} by at least 1.14.

\begin{figure*}[!t]
\footnotesize
\centering
    \begin{tabular}{c c c c c c c}
            \multicolumn{3}{c}{\multirow{5}*[45.6pt]{
            \hspace{-2.5mm} \includegraphics[width=0.325\linewidth,height=0.235\linewidth]{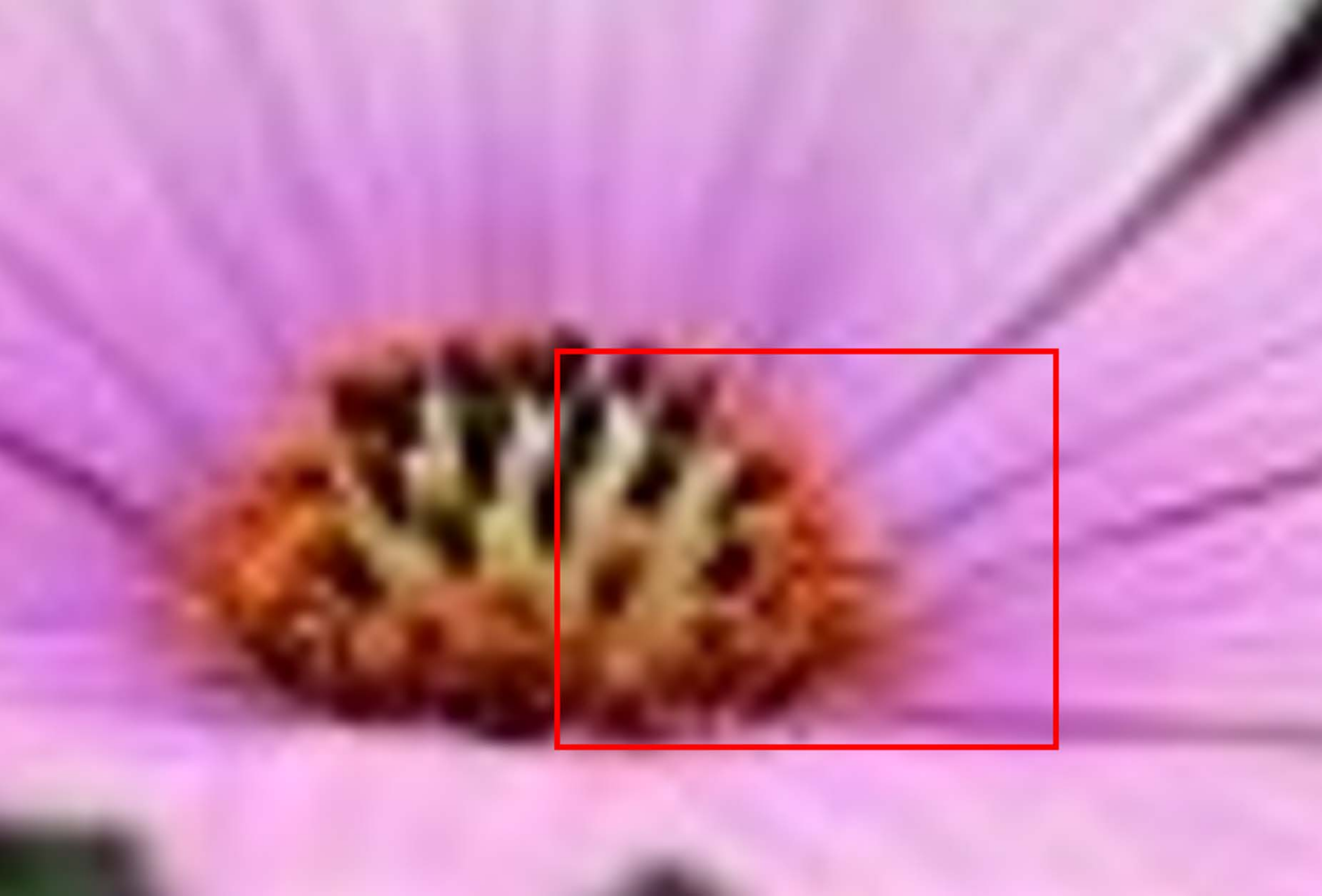}}}
            & \hspace{-4.0mm} \includegraphics[width=0.16\linewidth,height=0.105\linewidth]{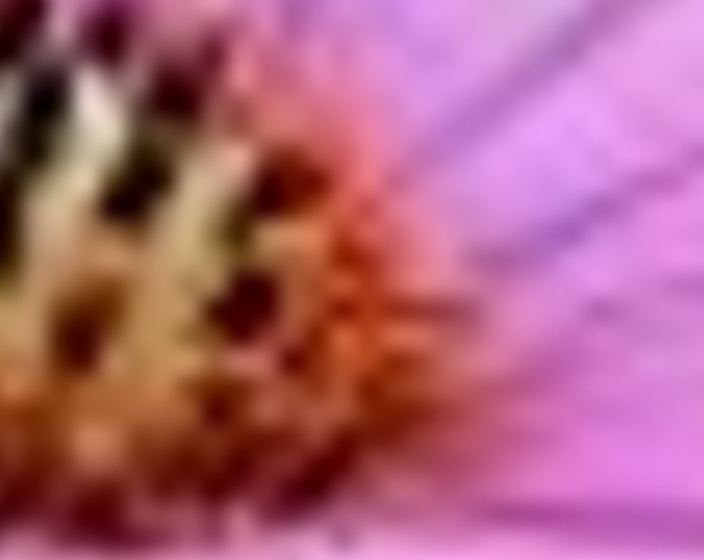}
            & \hspace{-4.0mm} \includegraphics[width=0.16\linewidth,height=0.105\linewidth]{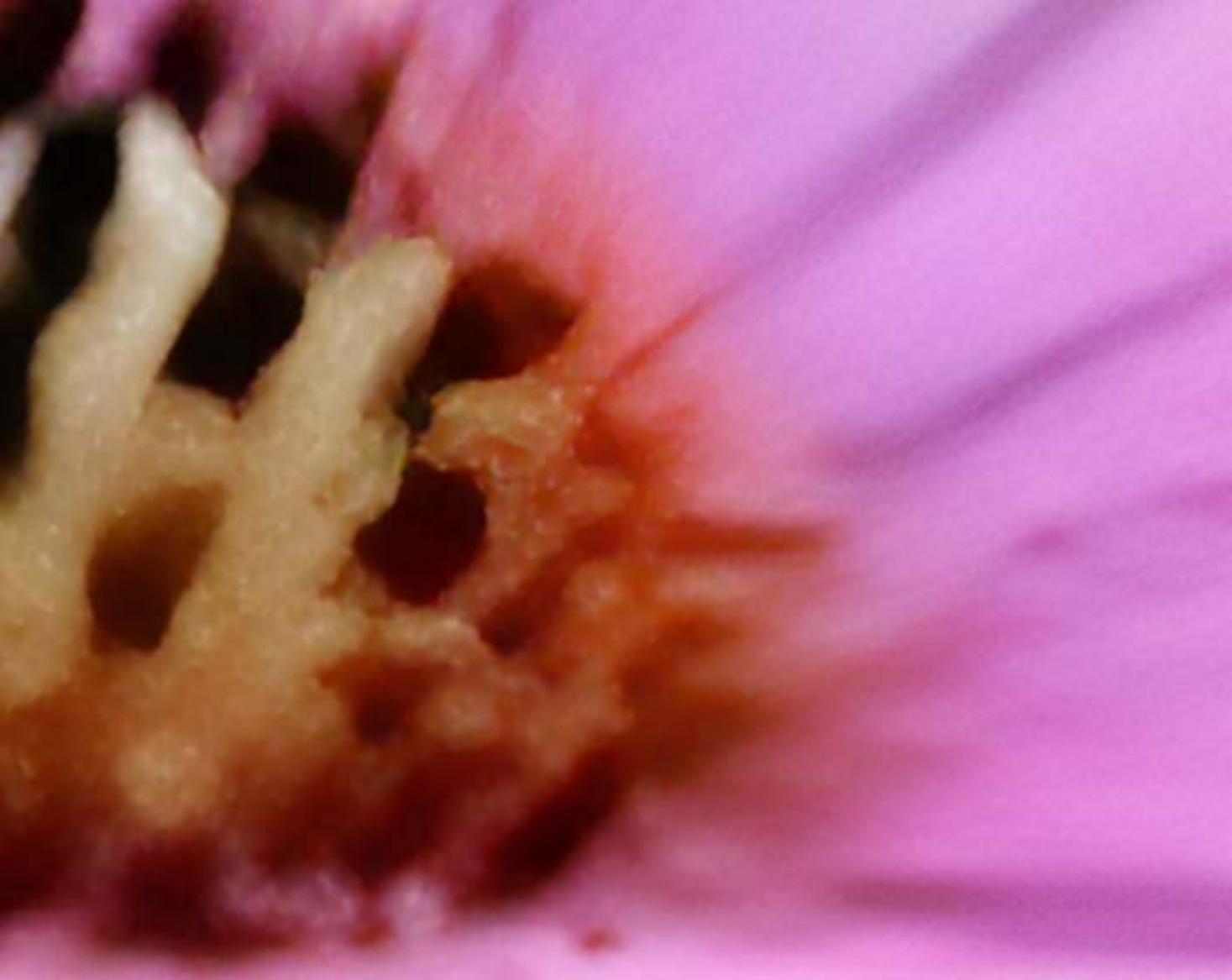}
            & \hspace{-4.0mm} \includegraphics[width=0.16\linewidth,height=0.105\linewidth]{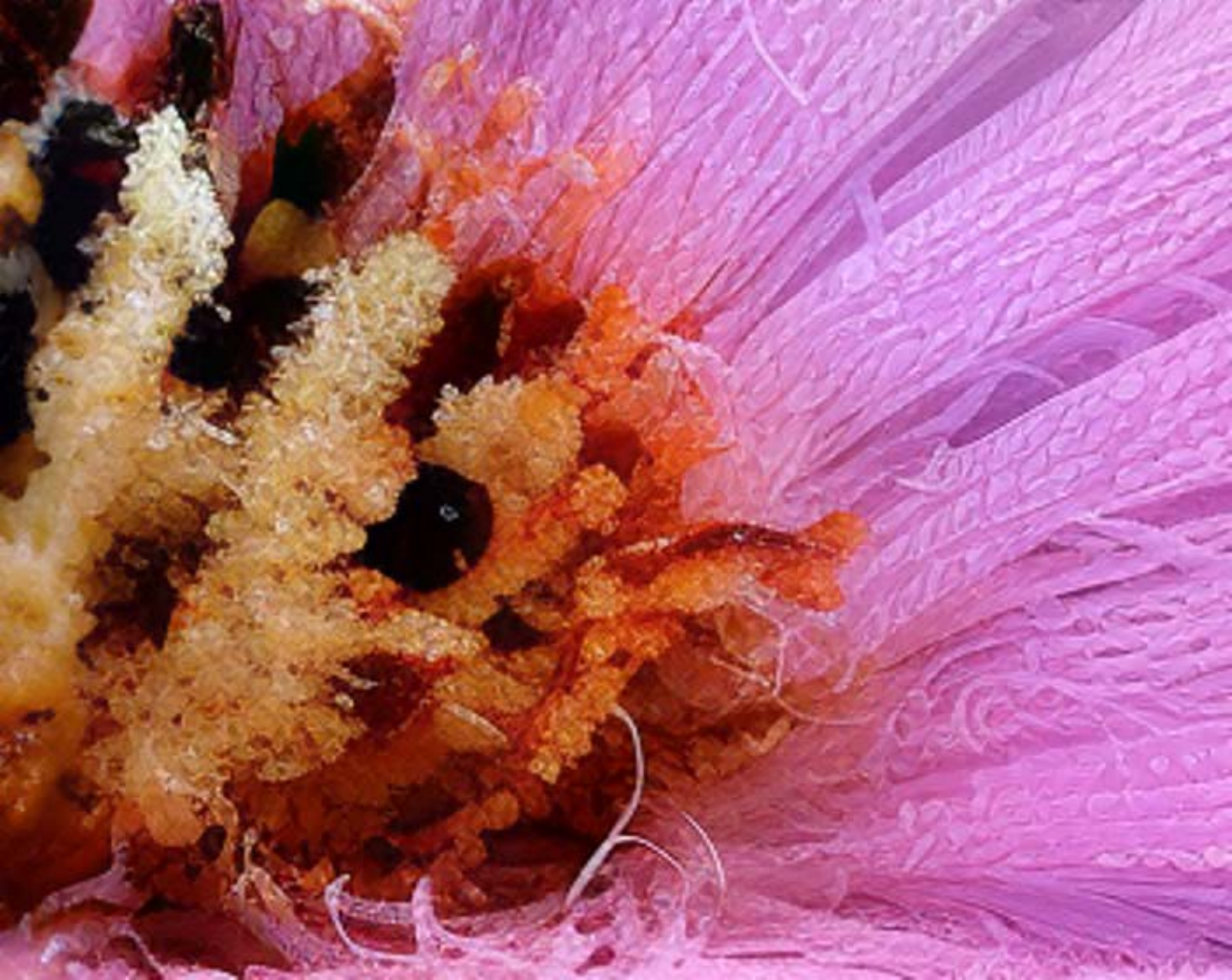}
            & \hspace{-4.0mm} \includegraphics[width=0.16\linewidth,height=0.105\linewidth]{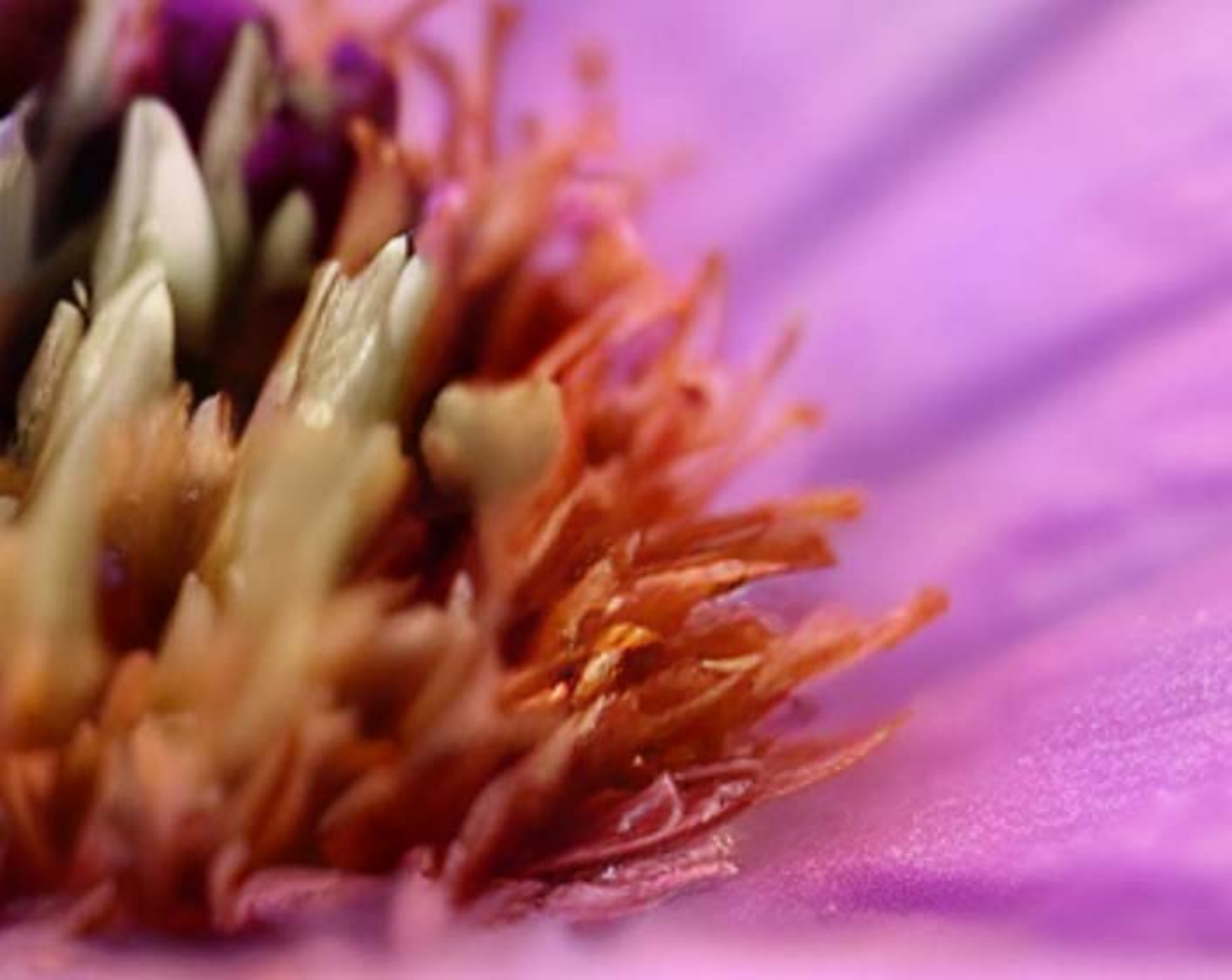}
              \\
    		\multicolumn{3}{c}{~}
            & \hspace{-4.0mm} (a) LQ patch
            & \hspace{-4.0mm} (b) Real-ESRGAN~\cite{Real-ESRGAN}
            & \hspace{-4.0mm} (c) DiffBIR~\cite{DiffBIR}
            & \hspace{-4.0mm} (d) PASD~\cite{PASD} \\		
    	\multicolumn{3}{c}{~}
            & \hspace{-4.0mm} \includegraphics[width=0.16\linewidth,height=0.105\linewidth]{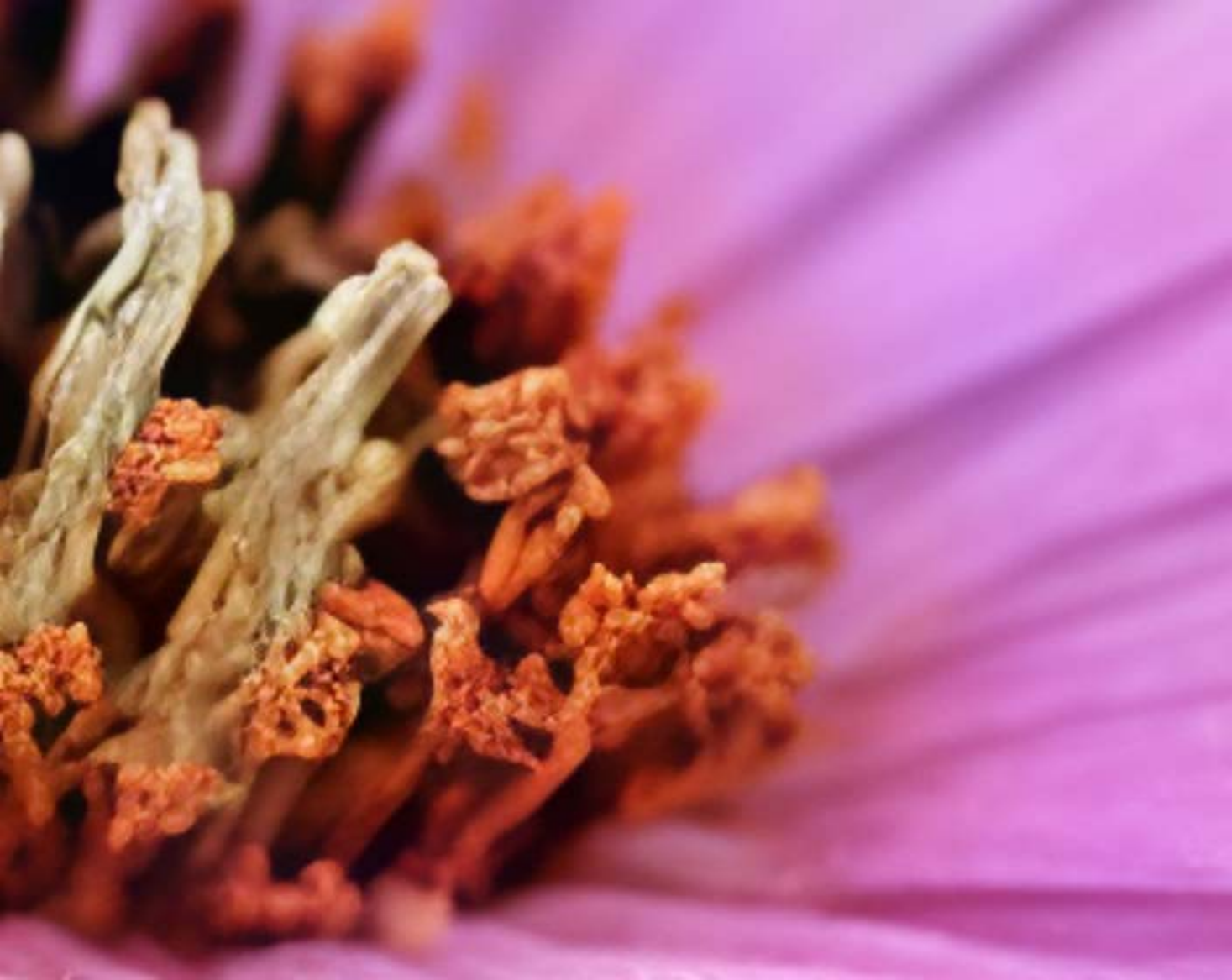}
            & \hspace{-4.0mm} \includegraphics[width=0.16\linewidth,height=0.105\linewidth]{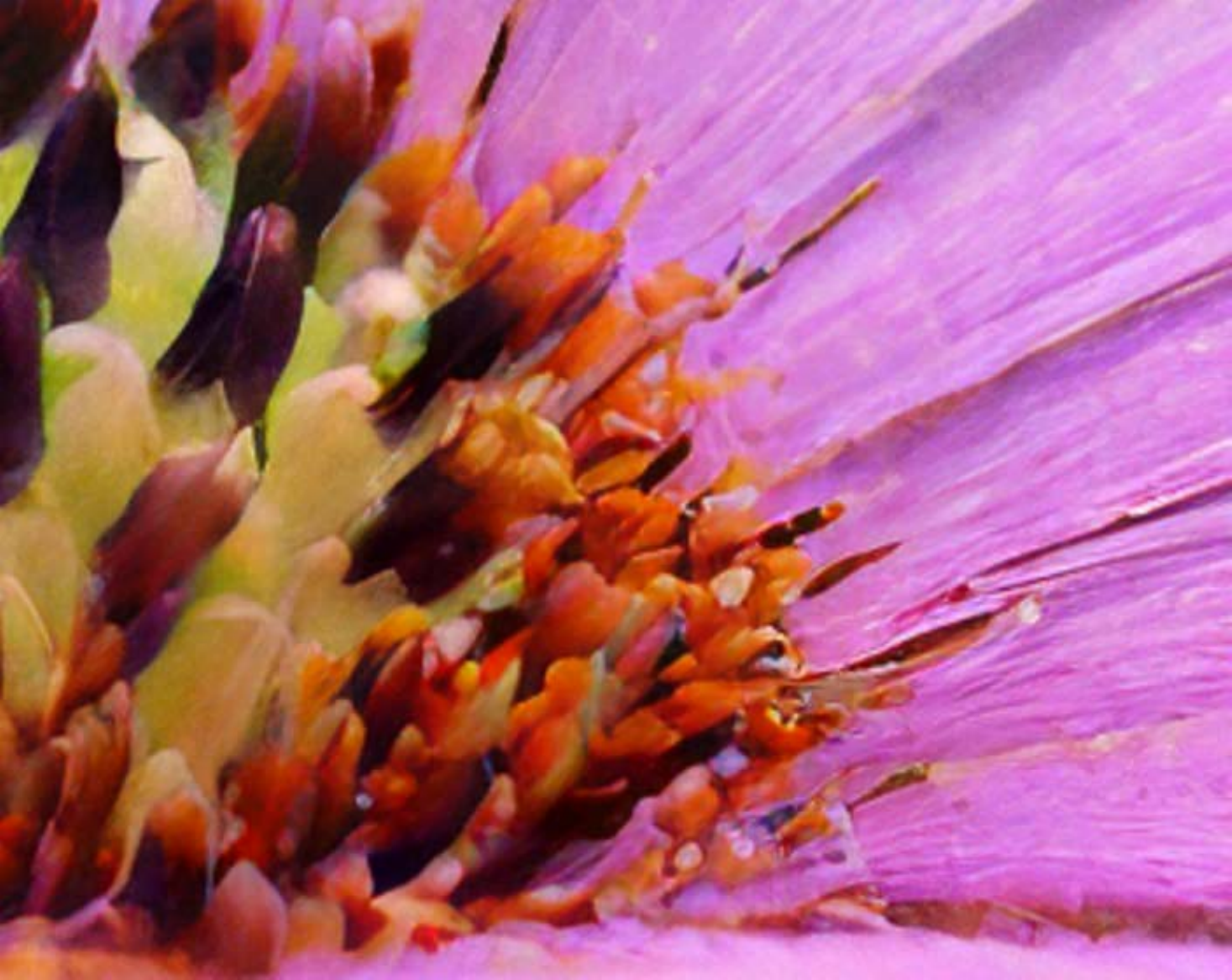}
            & \hspace{-4.0mm} \includegraphics[width=0.16\linewidth,height=0.105\linewidth]{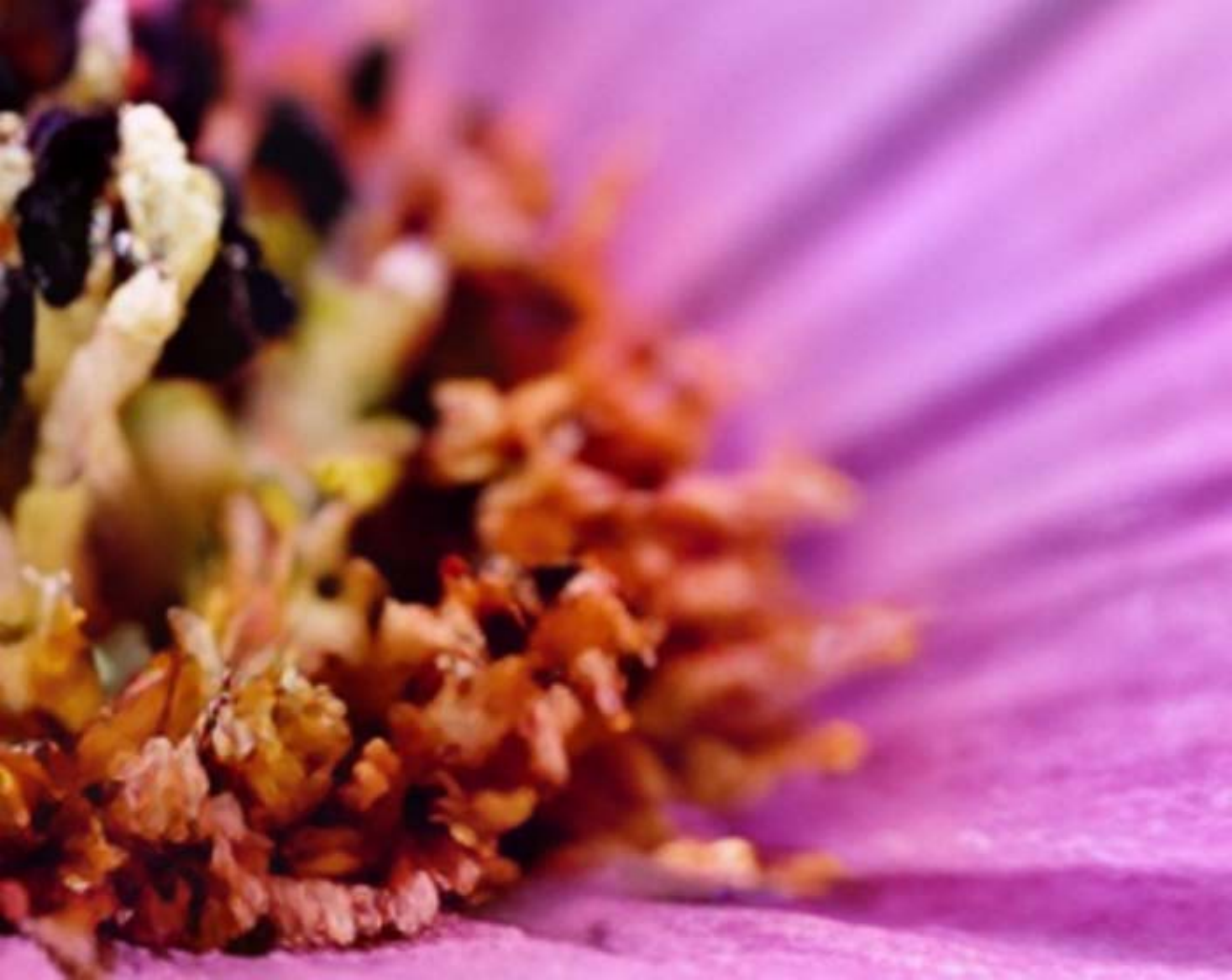}
            & \hspace{-4.0mm} \includegraphics[width=0.16\linewidth,height=0.105\linewidth]{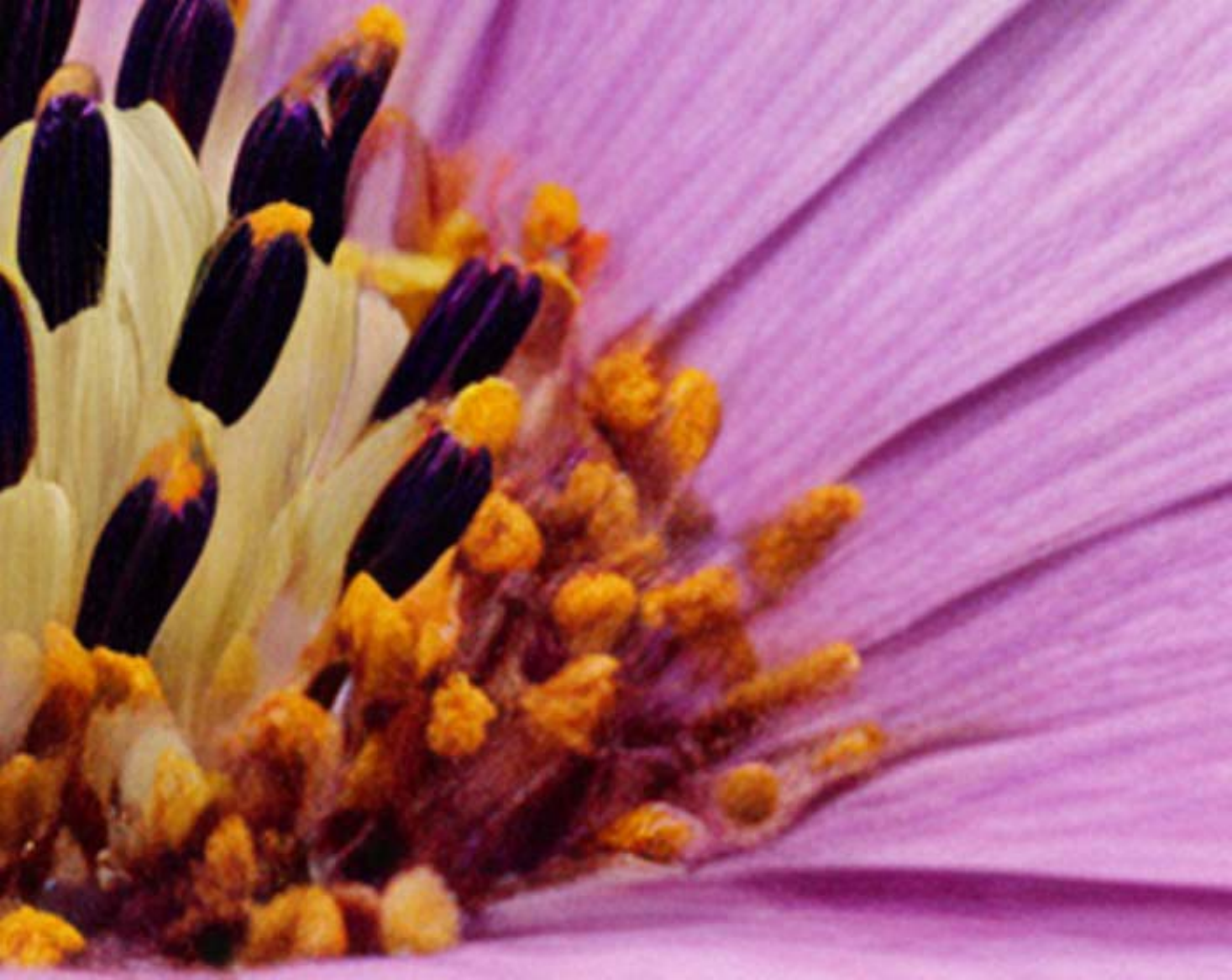}
            \\
    	\multicolumn{3}{c}{\hspace{-4.0mm} LQ image from RealPhoto60~\cite{SUPIR}}
            & \hspace{-4.0mm} (e) SeeSR~\cite{SeeSR}
            & \hspace{-4.0mm} (f) DreamClear~\cite{dreamclear}
            & \hspace{-4.0mm} (g) SUPIR~\cite{SUPIR}
            & \hspace{-4.0mm} (h) Ours\\

            \multicolumn{3}{c}{\multirow{5}*[45.6pt]{
            \hspace{-2.5mm} \includegraphics[width=0.325\linewidth,height=0.235\linewidth]{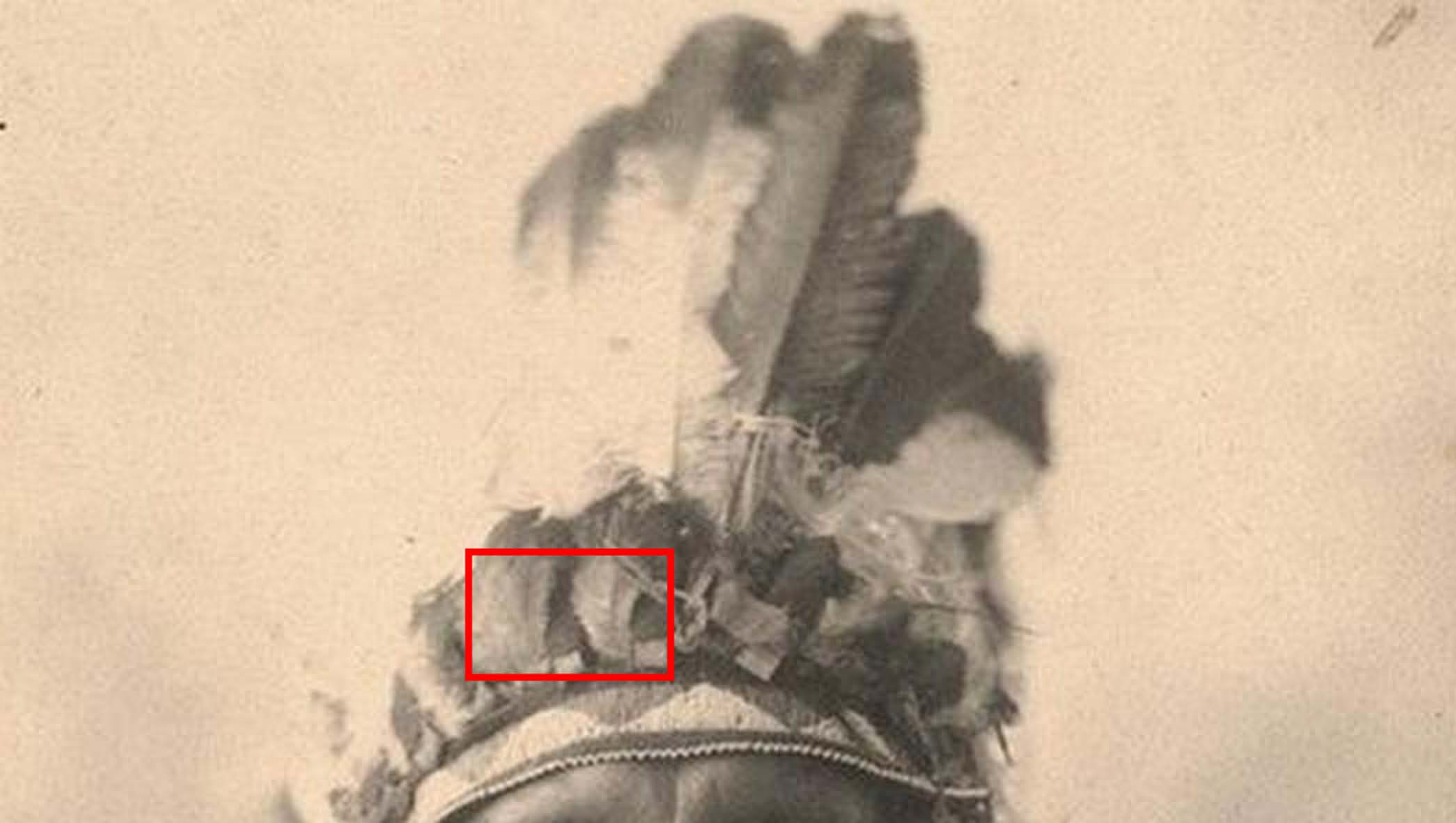}}}
            & \hspace{-4.0mm} \includegraphics[width=0.16\linewidth,height=0.105\linewidth]{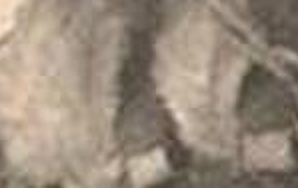}
            & \hspace{-4.0mm} \includegraphics[width=0.16\linewidth,height=0.105\linewidth]{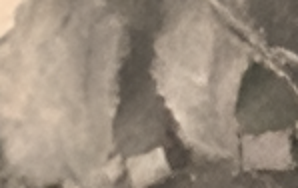}
            & \hspace{-4.0mm} \includegraphics[width=0.16\linewidth,height=0.105\linewidth]{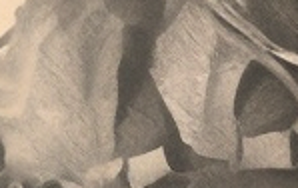}
            & \hspace{-4.0mm} \includegraphics[width=0.16\linewidth,height=0.105\linewidth]{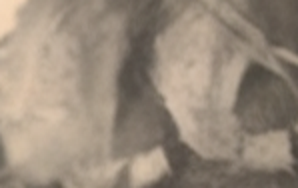}
              \\
    		\multicolumn{3}{c}{~}
            & \hspace{-4.0mm} (a) LQ patch
            & \hspace{-4.0mm} (b) Real-ESRGAN~\cite{Real-ESRGAN}
            & \hspace{-4.0mm} (c) DiffBIR~\cite{DiffBIR}
            & \hspace{-4.0mm} (d) PASD~\cite{PASD} \\		
    	\multicolumn{3}{c}{~}
            & \hspace{-4.0mm} \includegraphics[width=0.16\linewidth,height=0.105\linewidth]{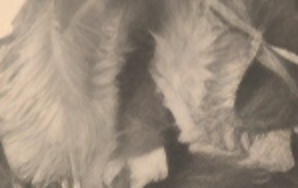}
            & \hspace{-4.0mm} \includegraphics[width=0.16\linewidth,height=0.105\linewidth]{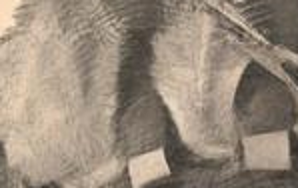}
            & \hspace{-4.0mm} \includegraphics[width=0.16\linewidth,height=0.105\linewidth]{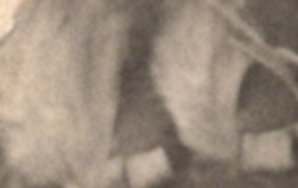}
            & \hspace{-4.0mm} \includegraphics[width=0.16\linewidth,height=0.105\linewidth]{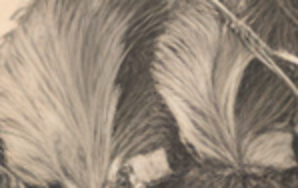}
            \\
    	\multicolumn{3}{c}{\hspace{-4.0mm} Old photo from RealDeg}
            & \hspace{-4.0mm} (e) SeeSR~\cite{SeeSR}
            & \hspace{-4.0mm} (f) DreamClear~\cite{dreamclear}
            & \hspace{-4.0mm} (g) SUPIR~\cite{SUPIR}
            & \hspace{-4.0mm} (h) Ours\\

            \multicolumn{3}{c}{\multirow{5}*[45.6pt]{
            \hspace{-2.5mm} \includegraphics[width=0.325\linewidth,height=0.235\linewidth]{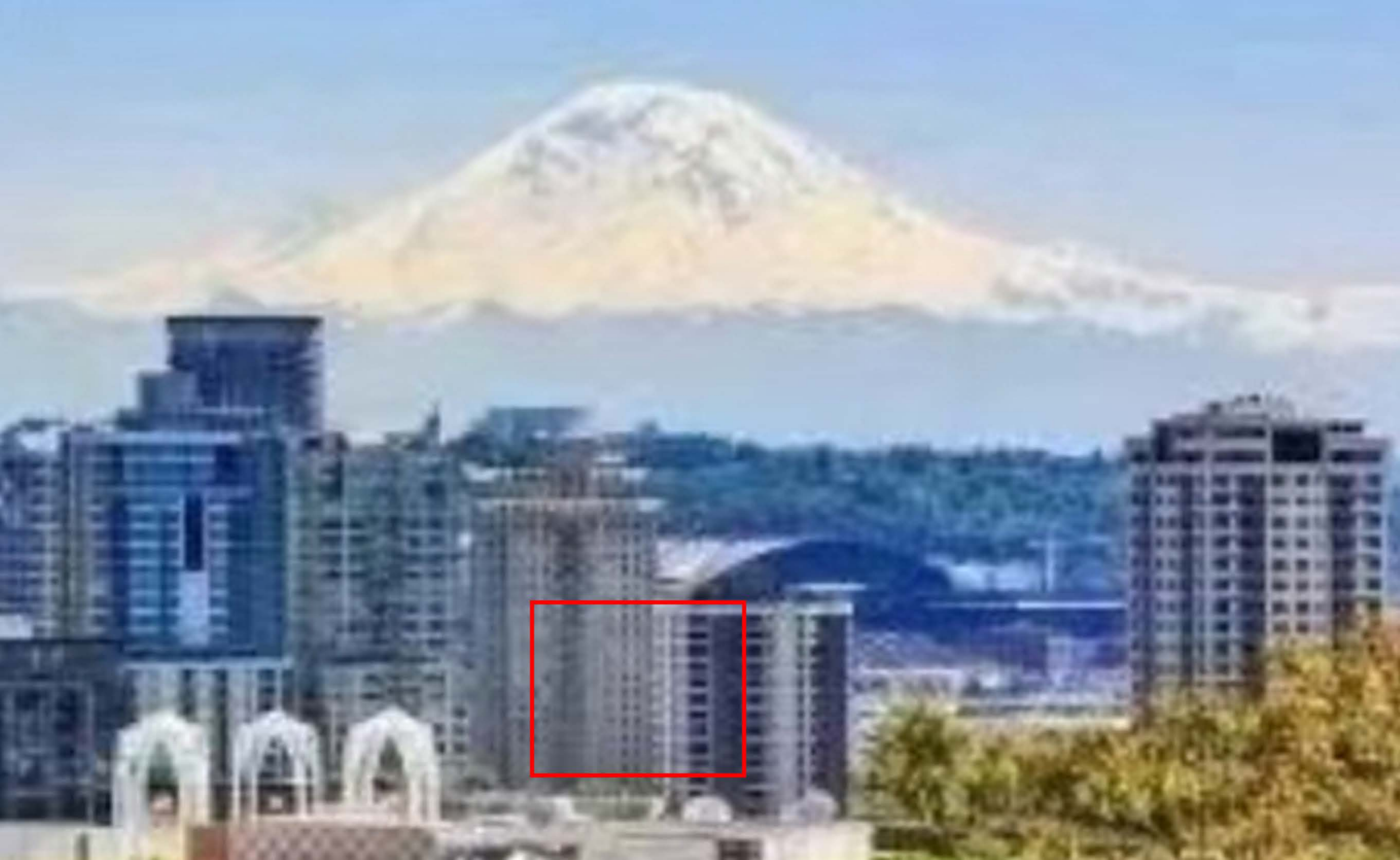}}}
            & \hspace{-4.0mm} \includegraphics[width=0.16\linewidth,height=0.105\linewidth]{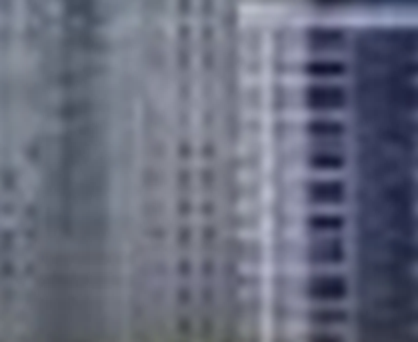}
            & \hspace{-4.0mm} \includegraphics[width=0.16\linewidth,height=0.105\linewidth]{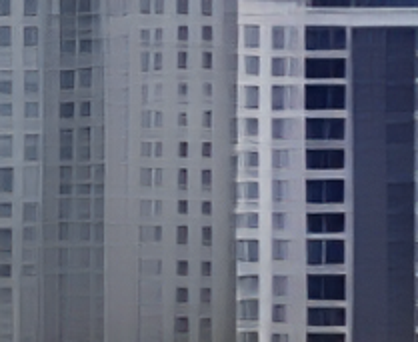}
            & \hspace{-4.0mm} \includegraphics[width=0.16\linewidth,height=0.105\linewidth]{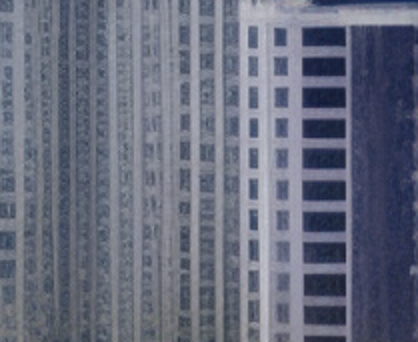}
            & \hspace{-4.0mm} \includegraphics[width=0.16\linewidth,height=0.105\linewidth]{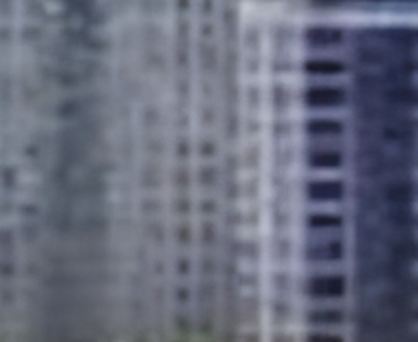}
              \\
    		\multicolumn{3}{c}{~}
            & \hspace{-4.0mm} (a) LQ patch
            & \hspace{-4.0mm} (b) Real-ESRGAN~\cite{Real-ESRGAN}
            & \hspace{-4.0mm} (c) DiffBIR~\cite{DiffBIR}
            & \hspace{-4.0mm} (d) PASD~\cite{PASD} \\		
    	\multicolumn{3}{c}{~}
            & \hspace{-4.0mm} \includegraphics[width=0.16\linewidth,height=0.105\linewidth]{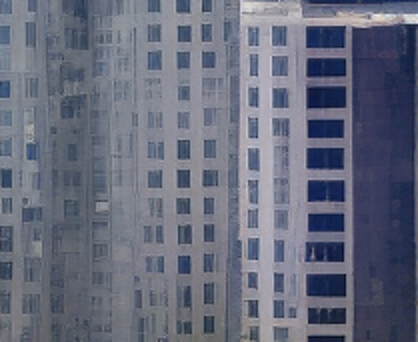}
            & \hspace{-4.0mm} \includegraphics[width=0.16\linewidth,height=0.105\linewidth]{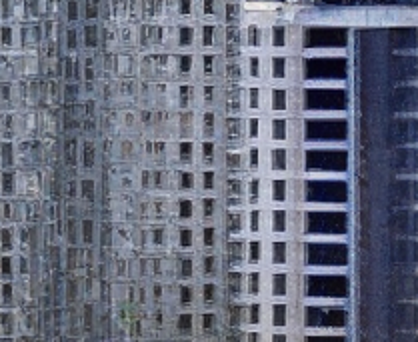}
            & \hspace{-4.0mm} \includegraphics[width=0.16\linewidth,height=0.105\linewidth]{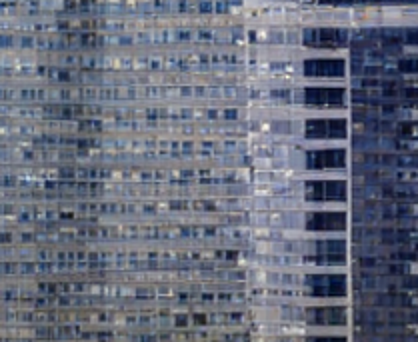}
            & \hspace{-4.0mm} \includegraphics[width=0.16\linewidth,height=0.105\linewidth]{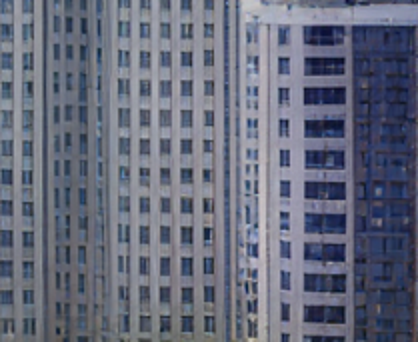}
            \\
    	\multicolumn{3}{c}{\hspace{-4.0mm} Social media image from RealDeg}
            & \hspace{-4.0mm} (e) SeeSR~\cite{SeeSR}
            & \hspace{-4.0mm} (f) DreamClear~\cite{dreamclear}
            & \hspace{-4.0mm} (g) SUPIR~\cite{SUPIR}
            & \hspace{-4.0mm} (h) Ours \\

    \end{tabular}
\vspace{-2mm}
\caption{Image SR result (×2) on the real-world benchmarks. Compared to competing methods, our approach generates more realistic images with fine-scale structures and details.}
\label{fig: vis_real_comp}
\vspace{-3mm}
\end{figure*}

Figure~\ref{fig: vis_real_comp} shows one example from RealPhoto60~\cite{SUPIR} and two examples from the RealDeg dataset.
The proposed method generates more realistic images with better fine-scale structures and details (Figure~\ref{fig: vis_real_comp}(h)).

\noindent \textbf{Run-time comparision.}
Benefiting from unleashing the diffusion prior and the alignment module, our approach does not need adaptors like ControlNet~\cite{controlnet} to steer the diffusion process with LQ features, which significantly improve our efficiency.
We further benchmark the run-time for the diffusion process of existing diffusion-based image SR methods and our approach on $10$ images with the size of 1024 $\times$ 1024 pixels in Table~\ref{tab:complexity}. All methods are tested on a machine with an A800 GPU. As shown in Tables~\ref{tab: synthetic_comp}-\ref{tab:complexity}, our proposed method is more efficient with better image quality.

\subsection{Real-world OCR recognition}
To further examine whether the proposed method generates faithful structures, we provide evaluation results on the OCR detection accuracy of the restored images.
We select 200 images from ICDAR 2024 Occluded RoadText dataset~\cite{ICDAR} and generate LQ images with severe degradations using the same method as described in Seciton~\ref{synthesis},
For OCR recognition, we use the robust visual backbone PaddleOCR v3~\cite{PPOCR} to evaluate on the restored images by various SR methods.
Table~\ref{tab: ocr} shows that the OCR recognition accuracy
of our restored images is better, indicating that our method is capable of effectively restoring reliable structural information.

\begin{table}[!t]
\small
\caption{Run-time performance of diffusion-based SR methods for the diffusion process. All methods are tested on a machine with an A800 GPU. The best and second performances are highlighted in {\color{red}red} and {\color{blue}blue}, respectively.}
\centering
\vspace{-2mm}
\resizebox{1.0\columnwidth}{!}{
\begin{tabular}{ccccccc}
\toprule
                  & DiffBIR~\cite{DiffBIR} & PASD~\cite{PASD} & SeeSR~\cite{SeeSR} & DreamClear~\cite{SeeSR} & SUPIR~\cite{SUPIR} & Ours \\ \bottomrule
Inference Step    & 50      & 20   & 50    &  50         & 50     & 20   \\
Running Time (s) & 46.81   & {\color{blue}7.31} & 10.31 &  7.58          & 11.44 & {\color{red}2.55}  \\ \bottomrule
\end{tabular}}
\vspace{-2mm}
\label{tab:complexity}
\end{table}


\section{Analysis and Discussion}
\label{sec: ablation_study}

\subsection{Effectiveness of the alignment module}
The alignment module is proposed to adaptively explore useful information from LQ features to facilitate the progressive diffusion process for faithful HQ image reconstruction.
To demonstrate the effectiveness of our alignment module, we compare with a baseline method that removes the alignment module (Ours$_{\text{w/o Align}}$ for short).
To leverage LQ features to guide the diffusion process, we respectively apply one convolutional layer on the noisy latent as well as the LQ features and then add their results together.
The quantitative results in Table~\ref{tab: fea_align} show that
the MUSIQ of our approach is 6.07 higher than the baseline method on the RealPhoto60 dataset.
The comparison results illustrate that the proposed alignment module is more effective in capturing useful information from LQ features to facilitate the diffusion process.

To further analyze the effect of pre-training the alignment module, we compare our approach with a baseline that does not pre-train the alignment module but directly trains it together with the whole network (Ours$_\text{w/o Pre-train align}$ for short).
The results in Table~\ref{tab: fea_align} demonstrate the effectiveness of pre-training the alignment module, which is more effective in exploiting useful features from LQ features and incorporating them with the noisy latant of the diffusion model, increasing the MUSIQ by 2.98 on the RealPhoto60 dataset~\cite{SUPIR}.

We additionally investigate whether it is necessary to use the visual features from the penultimate layer of the LQ encoder instead of those from the last layer.
To this end, we further compare with a baseline that replaces the LQ features with the visual features of the last layer of the LQ encoder (Ours$_{\text{w/ Last feats}}$ for short).
As Table~\ref{tab: fea_align} shows, our approach performs better than the baseline, decreasing the LPIPS by 0.0222 on the DIV2K-Val~\cite{DIV2K} dataset and increasing the MUSIQ by 2.69 on the RealPhoto60~\cite{SUPIR} dataset.
The comparison results demonstrate that the features extracted from the penultimate layer of the VAE encoder can effectively capture more useful information.

\setcounter{figure}{5}
\begin{figure*}[!t]
\scriptsize
\centering
\begin{tabular}{@{}c@{\hspace{1mm}}c@{\hspace{1mm}}c@{\hspace{1mm}}c@{\hspace{1mm}}c@{\hspace{1mm}}c@{\hspace{1mm}}c@{}c@{\hspace{1mm}}c@{}}
\includegraphics[width = 0.138\linewidth, ,height=0.115\linewidth]{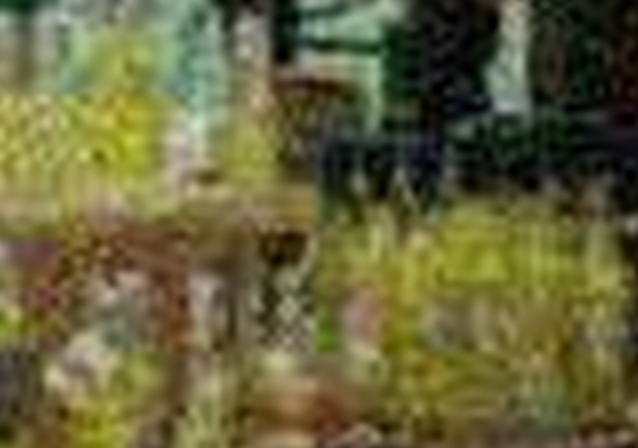}&
\includegraphics[width = 0.138\linewidth,height=0.115\linewidth]{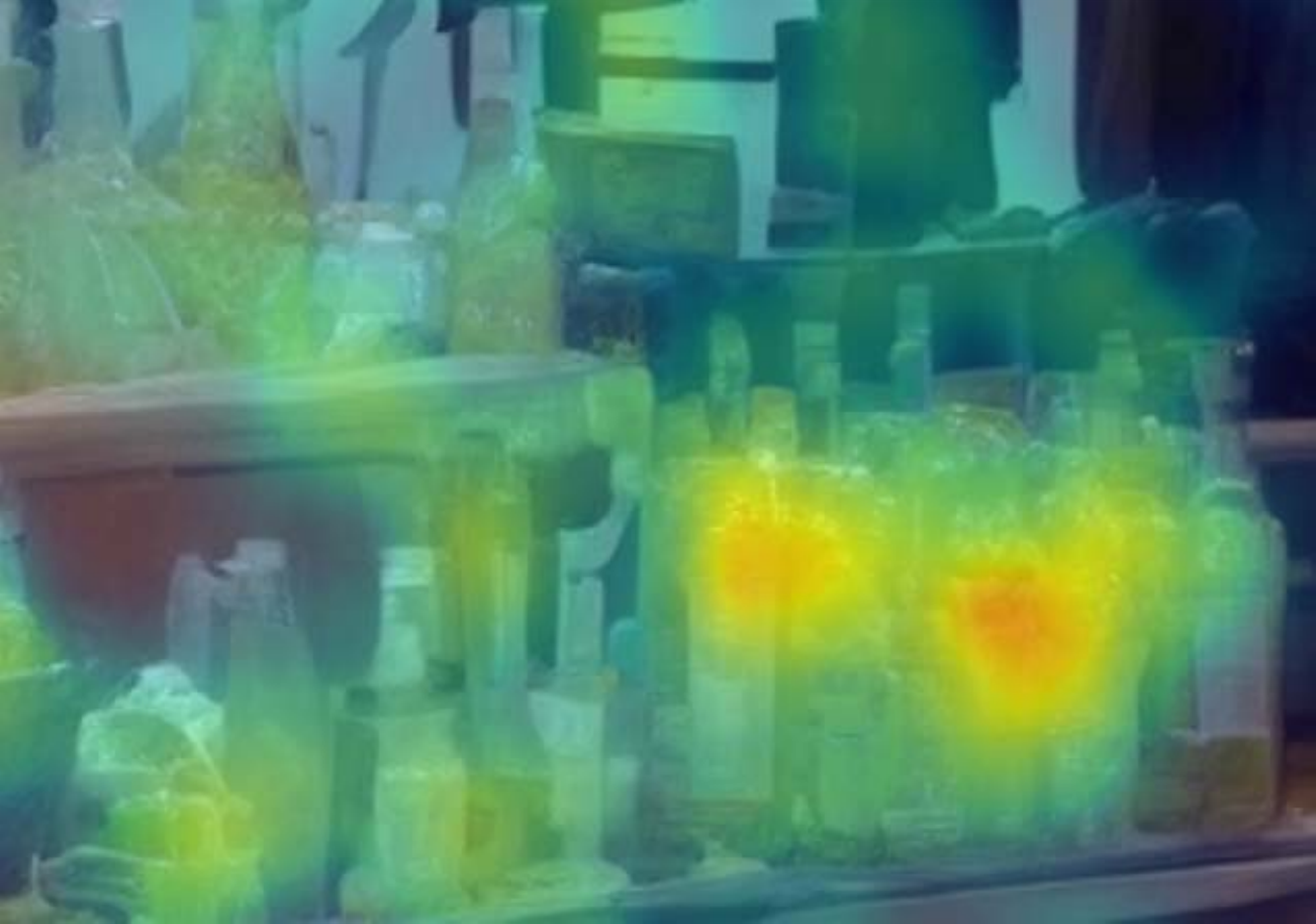}&
\includegraphics[width = 0.138\linewidth,height=0.115\linewidth]{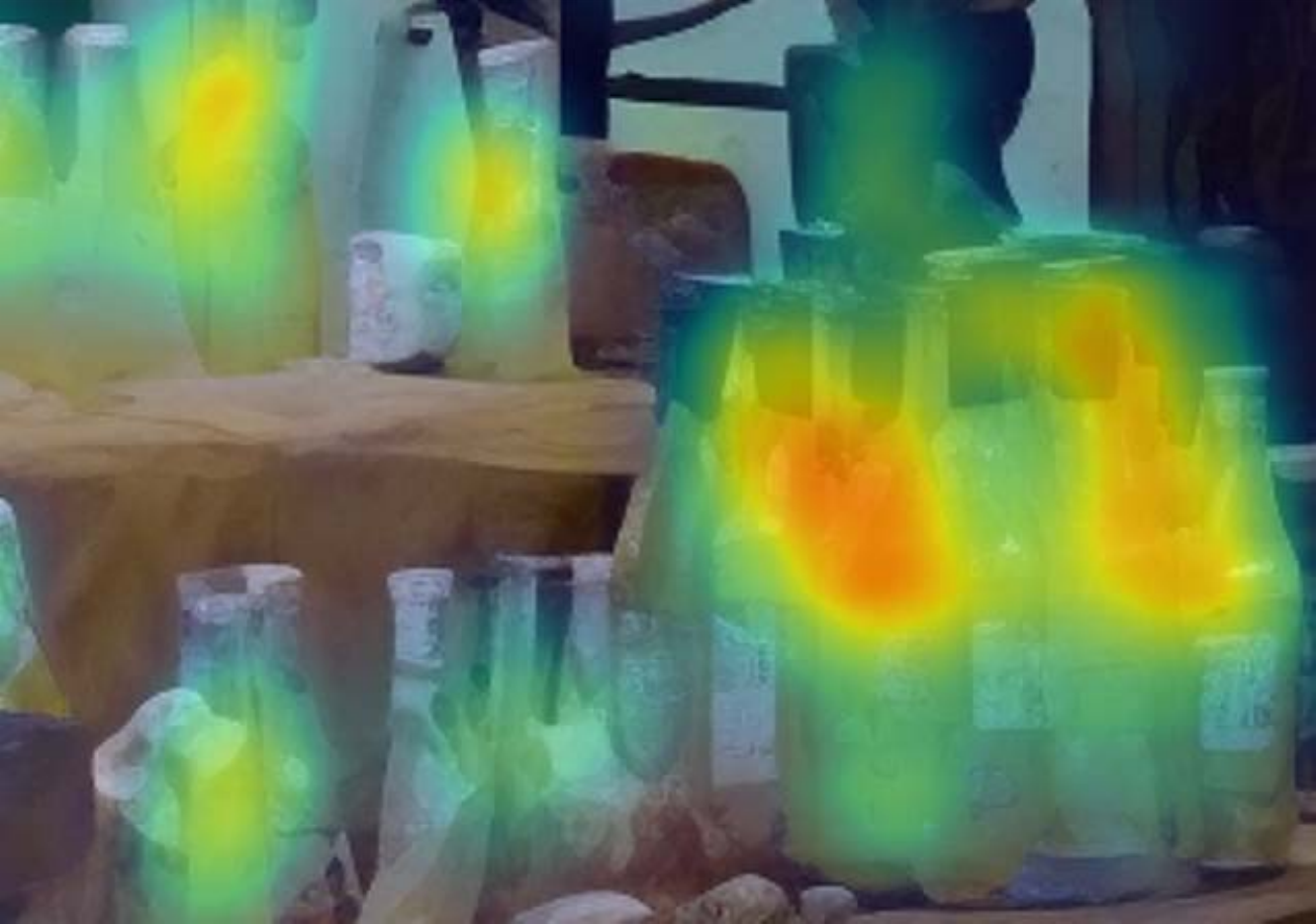}&
\includegraphics[width = 0.138\linewidth,height=0.115\linewidth]{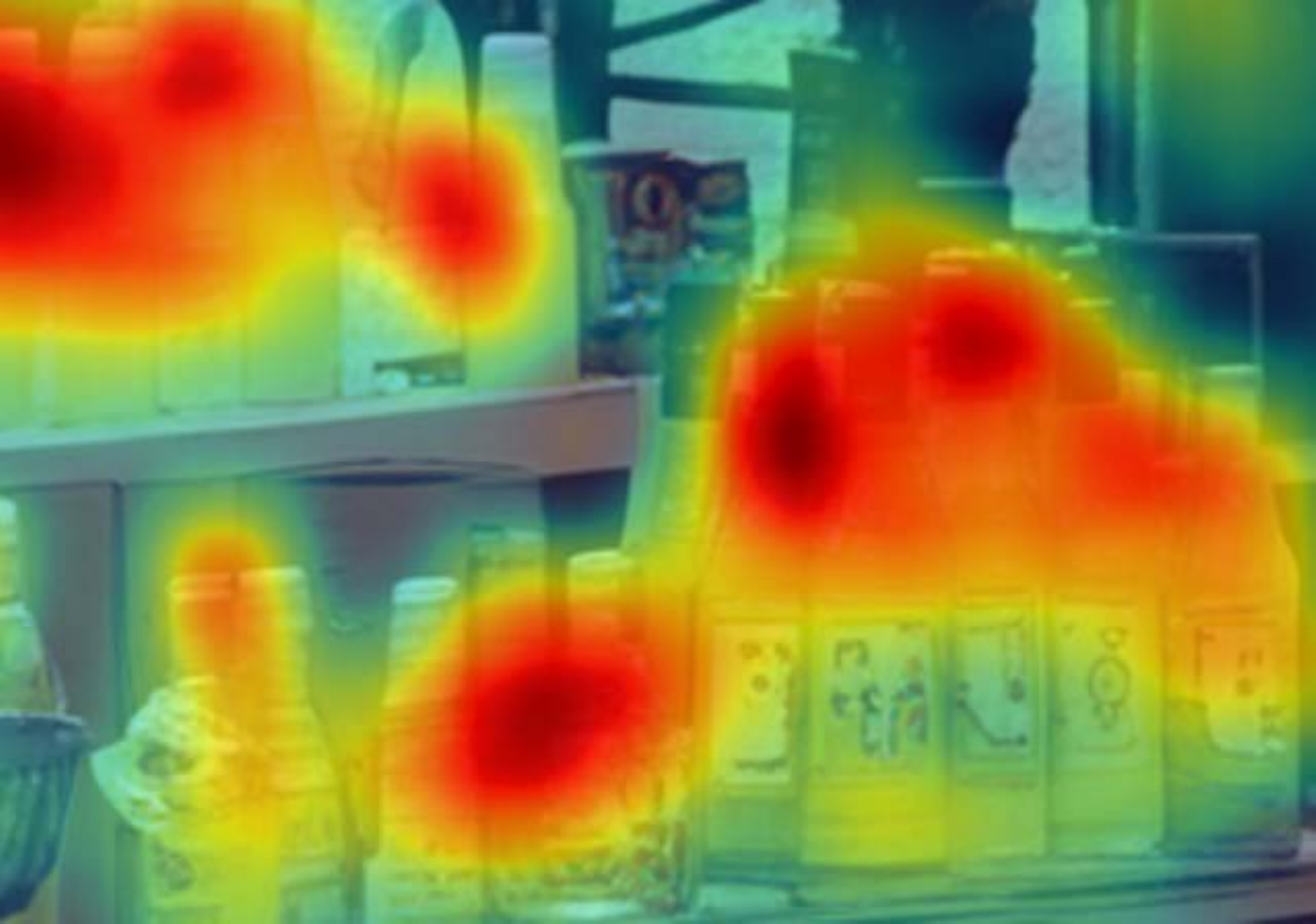}&
\includegraphics[width = 0.138\linewidth,height=0.115\linewidth]{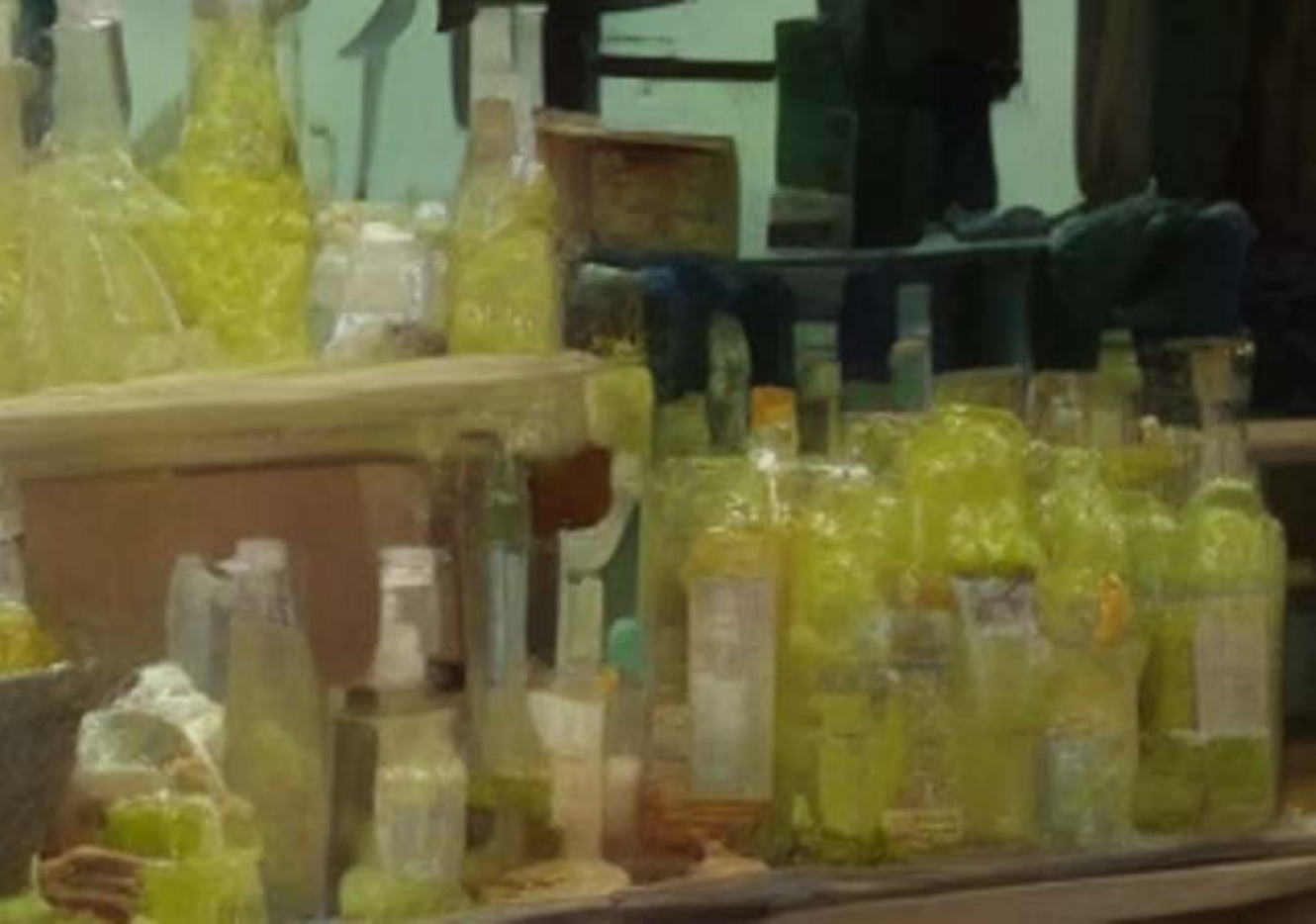}&
\includegraphics[width = 0.138\linewidth,height=0.115\linewidth]{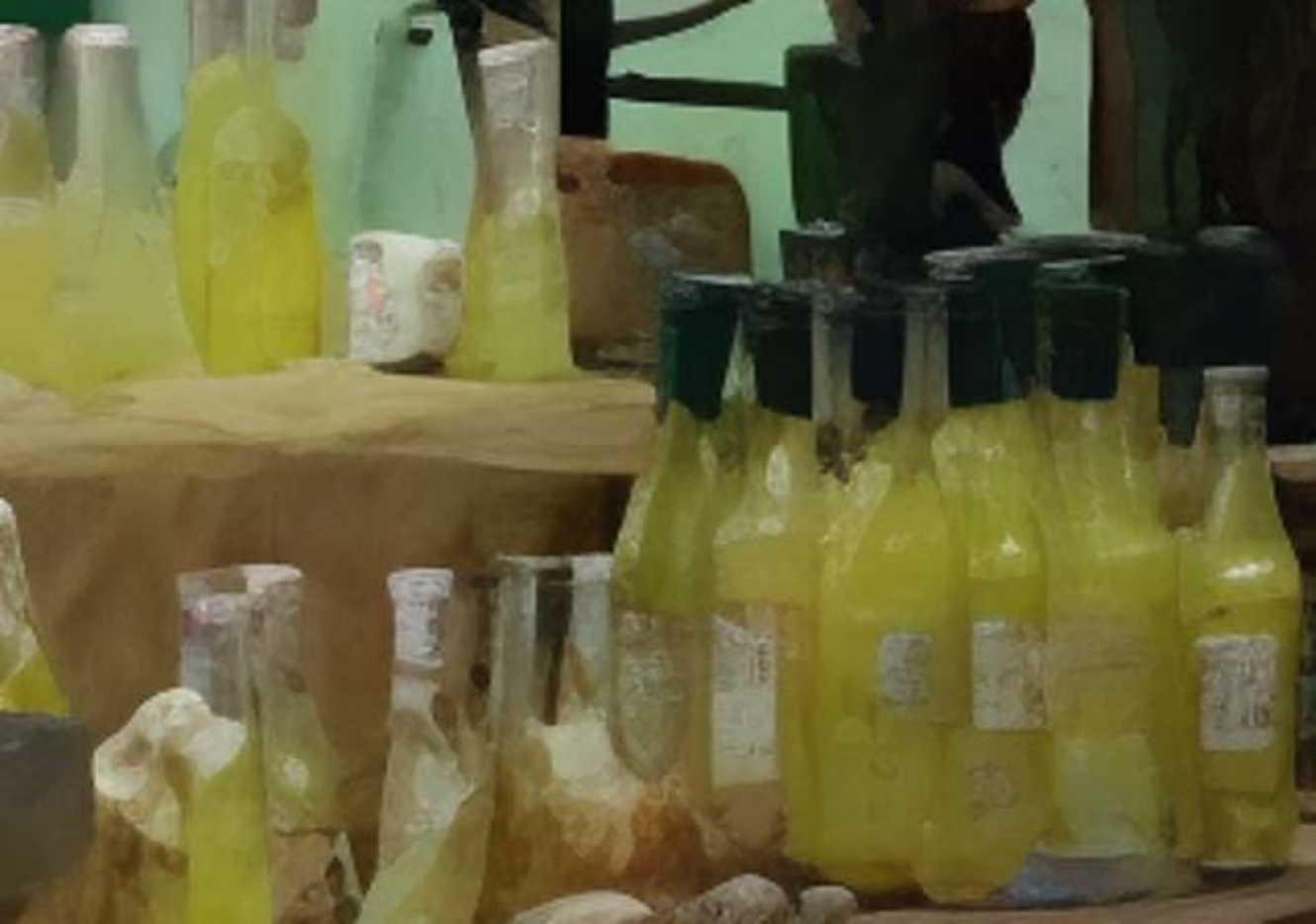}&
\includegraphics[width = 0.138\linewidth,height=0.115\linewidth]{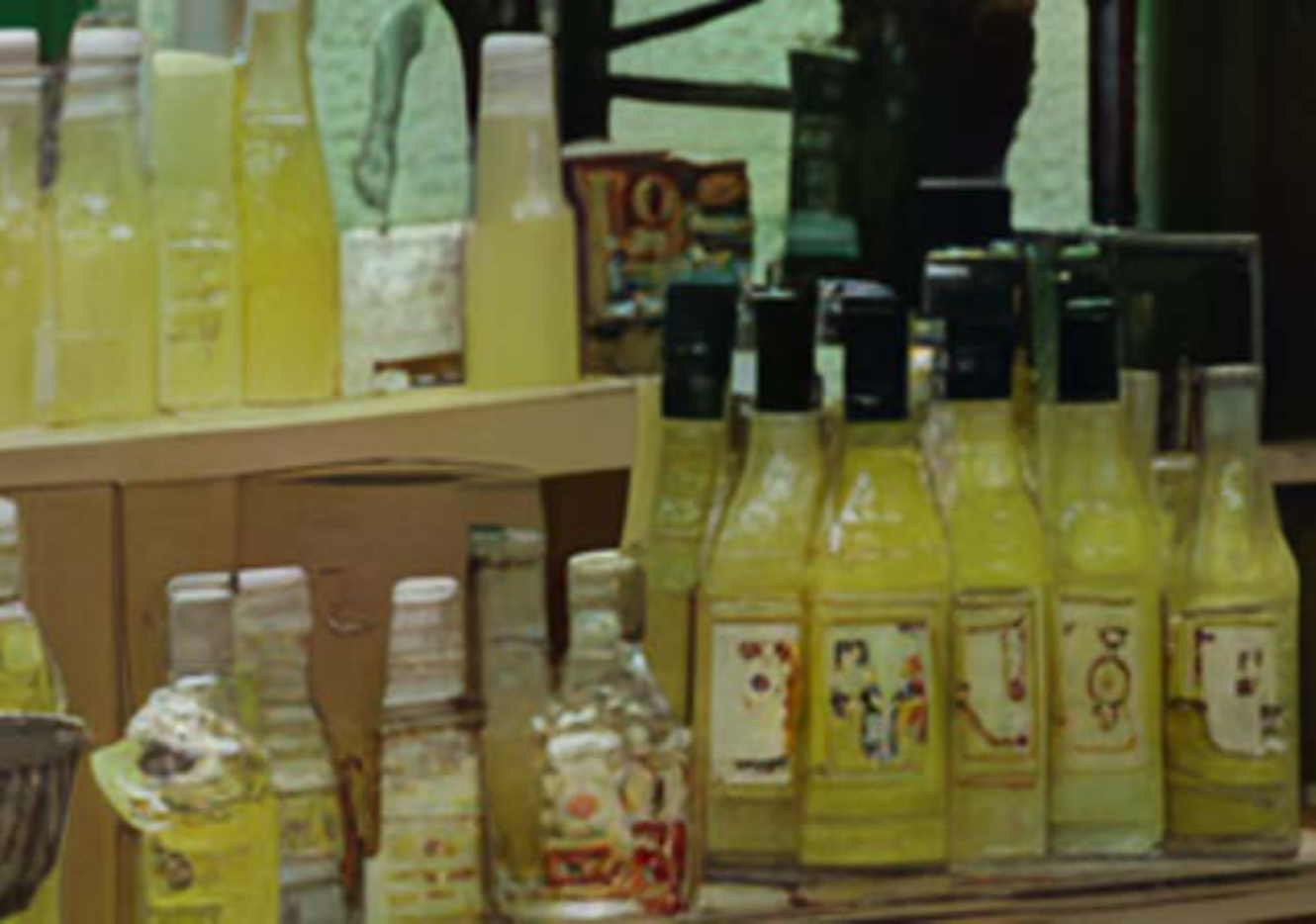} \\
(a) LQ Patch & (b) & (c) & (d) & (e) PASD~\cite{PASD} & (f) SeeSR~\cite{SeeSR} & (g) Ours \\
\end{tabular}
\vspace{-4mm}
\caption{Visualization of DAAMs~\cite{DAAM} for `bottles'. (b)-(d) are DAAMs for PASD~\cite{PASD}, SeeSR~\cite{SeeSR}, and our method.}
\label{fig: ablation_bottle}
\vspace{-6mm}
\end{figure*}

\subsection{Effectiveness of unified feature optimization}
To demonstrate the effectiveness of the proposed unified feature optimization strategy, we compare with baseline methods that respectively fix the diffusion model (FT EN \& Fix DM for short), fix the encoder (Fix EN \& FT DM for short), as well as separately fine-tune the encoder and diffusion model (FT EN \& DM (SP) for short).
Note that for FT EN \& Fix DM, we use ControlNet~\cite{controlnet} with the spatial feature transfer block (SFT)~\cite{SUPIR} to make the LQ features suitable for guiding the diffusion process.
As shown in Table~\ref{tab: unified_fea}, the proposed approach outperforms all baseline methods, increasing the MUSIQ by at least 1.63 on the RealPhoto60~\cite{SUPIR} dataset.
This demonstrates the importance of jointly optimizing the encoder and the diffusion model, which can benefit from their interplay.

\begin{table}[!t]
\caption{Effectiveness of the proposed alignment module. All methods are trained using the same settings as the proposed method for fair comparison.}
\vspace{-2mm}
\centering
\resizebox{1.0\columnwidth}{!}{
\begin{tabular}{lccccc}
\toprule
\multicolumn{1}{c}{} &
  \multirow{2}{*}{\begin{tabular}[c]{@{}c@{}}Alignment\\ module\end{tabular}} &
  \multirow{2}{*}{\begin{tabular}[c]{@{}c@{}}Pre-train\\ alignment\end{tabular}} &
  \multirow{2}{*}{\begin{tabular}[c]{@{}c@{}}Penultimate\\ visual features\end{tabular}} &
  DIV2K-Val~\cite{DIV2K} &
  RealPhoto60~\cite{SUPIR} \\ \cline{5-6}
\multicolumn{1}{c}{} &                             &                             &                             & LPIPS~{\color{red}$\downarrow$} & MUSIQ~{\color{red}$\uparrow$}  \\ \hline
$\text{Ours}_\text{w/o Align}$                 & \XSolidBrush & \Checkmark   & \Checkmark   & 0.3199      & 66.67              \\
$\text{Ours}_\text{w/o Pre-train align}$                 & \Checkmark   & \XSolidBrush & \Checkmark   & 0.3244      & 69.76             \\
$\text{Ours}_\text{w/ Last feats}$                 & \Checkmark   & \Checkmark   & \XSolidBrush & 0.3302      & 70.05             \\
Ours                 & \Checkmark   & \Checkmark   & \Checkmark   & 0.3080      & 72.74           \\ \bottomrule
\end{tabular}}
\vspace{-2mm}
\label{tab: fea_align}
\end{table}

\begin{table}[!t]
\caption{Effectiveness of unified feature optimization. All methods are trained using the same settings for fair comparison. SP denotes separately fine-tuning encoder and diffusion model.}
\centering
\vspace{-2mm}
\resizebox{1.0\columnwidth}{!}{
\begin{tabular}{lcccccccc}
\toprule
     & \multicolumn{2}{c}{Encoder (EN)}            &             & \multicolumn{2}{c}{Diffusion model (DM)}         &              & DIV2K-Val~\cite{DIV2K} & RealPhoto60~\cite{SUPIR} \\ \cline{2-3} \cline{5-6} \cline{8-9}
     & Fix                         & Fine-tune (FT)       &             & Fix                         & Fine-tune (FT)       &                     & LPIPS~{\color{red}$\downarrow$}       & MUSIQ~{\color{red}$\uparrow$}  \\ \hline
$\text{FT EN \& Fix DM}$  & \XSolidBrush & \Checkmark  & & \Checkmark  & \XSolidBrush &  & 0.3370 & 69.66        \\
$\text{Fix EN \& FT DM}$  & \Checkmark   & \XSolidBrush & & \XSolidBrush & \Checkmark &  & 0.3302 & 71.11        \\
$\text{FT EN \& DM (SP)}$ & \XSolidBrush & \Checkmark  & & \XSolidBrush & \Checkmark  &  & 0.3261 & 69.94        \\
Ours & \XSolidBrush & \Checkmark  & & \XSolidBrush & \Checkmark  &  & 0.3080 & 72.74      \\ \bottomrule
\end{tabular}}
\vspace{-6mm}
\label{tab: unified_fea}
\end{table}

Figure~\ref{fig:ablation_vis_comp} shows a visual comparison on an image from DIV2K-Val~\cite{DIV2K} with severe degradations.
The main structures generated by FT EN \& Fix DM are inconsistent with the ground truth, cf. the `windows' obtained by FT EN \& Fix DM in Figure~\ref{fig:ablation_vis_comp}(b) vs. the characters `MAER' of the GT in Figure~\ref{fig:ablation_vis_comp}(d).
As the diffusion model is pre-trained on HQ images, any mistakes in the LQ features can mislead the diffusion model into generating incorrect image structures.
The comparison results in Figure~\ref{fig:ablation_vis_comp} show that unleashing the diffusion model is able to improve the performance, but only yields faithful results when jointly optimized with the encoder.
This demonstrates that the proposed unified feature optimization strategy is able to benefit from the interplay of the encoder and the diffusion model, resulting in high-fidelity results, as shown in Figure~\ref{fig:ablation_vis_comp}(f).



\subsection{Visualization of DAAMs}
We visualize diffusion attentive attribution maps (DAAMs) \cite{DAAM} of some state-of-the-art diffusion-based methods and our approach in Figure~\ref{fig: ablation_bottle}.
Given an LQ image and the text `bottle' from the prompt `An image of an outdoor market with a lot of food and bottles', we upscale and aggregate cross-attention maps
in LDM during diffusion process to fetch DAAMs~\cite{DAAM}.
For LQ images with severe degradation in Figure~\ref{fig: ablation_bottle}(a), it is hard to extract accurate structural information, thus existing  ControlNet-based methods~\cite{PASD, SeeSR} yield weaker responses when calculating the similarity between LQ features and text embeddings (Figure~\ref{fig: ablation_bottle}(b)-(c)).
This hinders the activation of pre-trained diffusion priors and the restoration of faithful results as shown in Figure~\ref{fig: ablation_bottle}(e)-(f).
In contrast, our proposed approach unleashes diffusion priors and jointly optimizes the encoder and diffusion model to enable them to benefit from each other.
Thus, our method can effectively capture more useful information from LQ images, leading to a more plausible DAAM in Figure~\ref{fig: ablation_bottle}(d) and more faithful results in Figure~\ref{fig: ablation_bottle}(g).

\setcounter{figure}{4}
\begin{figure}[!t]
\scriptsize
\vspace{1mm}
\centering
\begin{tabular}
{@{}c@{\hspace{2mm}}c@{\hspace{2mm}}c@{\hspace{2mm}}c@{}}
\includegraphics[width = 0.315\linewidth, height=0.135\linewidth]{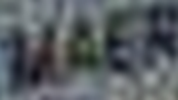}&
\includegraphics[width = 0.315\linewidth, height=0.135\linewidth]{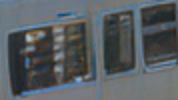}&
\includegraphics[width = 0.315\linewidth, height=0.135\linewidth]{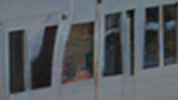} \\
(a) LQ Patch &\hspace{-3mm}
(b) FT EN \text{\&} Fix DM  &\hspace{-3mm}
(c) Fix EN \text{\&} FT DM\\
\includegraphics[width = 0.315\linewidth, height=0.135\linewidth]{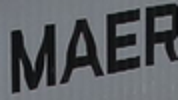}&
\includegraphics[width = 0.315\linewidth, height=0.135\linewidth]{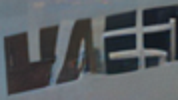}&
\includegraphics[width = 0.315\linewidth, height=0.135\linewidth]{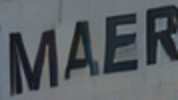} \\
(d) GT Patch &\hspace{-3mm}
(e) FT EN \text{\&} DM (SP) &\hspace{-3mm}
(f) Ours \\
\end{tabular}
\vspace{-3mm}
\caption{Effectiveness of the unified feature optimization on image SR ($\times 4$). Using unify optimization strategy is able to generate the results with clearer structural details.}
\label{fig:ablation_vis_comp}
\vspace{-7mm}
\end{figure}


\section{Conclusion}

In this paper, we propose an effective image SR method, named FaithDiff, which unleashes diffusion priors to better harness the powerful representation ability of LDM.
We show that unleashing diffusion priors is more effective in exploring useful information from degraded inputs and restoring faithful results.
We further develop an alignment module that effectively incorporates the features extracted by the encoder with the noisy latent of the diffusion model.
Taking all the components into one trainable network, we jointly optimize the encoder, alignment module, and the diffusion model.
This unified optimization strategy enables the encoder to provide useful features that coincide with the diffusion process, while the latent diffusion model can further restore these features.
Benefiting from their interplay, the entire network can effectively distinguish between degradation effects and inherent structural information, resulting in high-quality restoration.
Extensive experiments on synthetic and real-world benchmarks demonstrate that FaithDiff outperforms state-of-the-art methods in terms of structural fidelity and visual quality.
%

\clearpage

{
    \small
    \bibliographystyle{ieeenat_fullname}
    \bibliography{main}
}

\end{document}


%% file: arxiv.bbl
\begin{thebibliography}{49}
\providecommand{\natexlab}[1]{#1}
\providecommand{\url}[1]{\texttt{#1}}
\expandafter\ifx\csname urlstyle\endcsname\relax
  \providecommand{\doi}[1]{doi: #1}\else
  \providecommand{\doi}{doi: \begingroup \urlstyle{rm}\Url}\fi

\bibitem[Agustsson and Timofte(2017)]{DIV2K}
Eirikur Agustsson and Radu Timofte.
\newblock Ntire 2017 challenge on single image super-resolution: Dataset and
  study.
\newblock In \emph{CVPR Workshops}, 2017.

\bibitem[Ai et~al.(2024)Ai, Zhou, Huang, Han, Chen, You, and Yang]{dreamclear}
Yuang Ai, Xiaoqiang Zhou, Huaibo Huang, Xiaotian Han, Zhengyu Chen, Quanzeng
  You, and Hongxia Yang.
\newblock Dreamclear: High-capacity real-world image restoration with
  privacy-safe dataset curation.
\newblock In \emph{NeurIPS}, 2024.

\bibitem[Chan et~al.(2021)Chan, Wang, Xu, Gu, and Loy]{GLEAN}
Kelvin~CK Chan, Xintao Wang, Xiangyu Xu, Jinwei Gu, and Chen~Change Loy.
\newblock Glean: Generative latent bank for large-factor image
  super-resolution.
\newblock In \emph{CVPR}, 2021.

\bibitem[Chen et~al.(2022)Chen, Shi, Qin, Li, Han, Yang, and Guo]{FeMaSR}
Chaofeng Chen, Xinyu Shi, Yipeng Qin, Xiaoming Li, Xiaoguang Han, Tao Yang, and
  Shihui Guo.
\newblock Real-world blind super-resolution via feature matching with implicit
  high-resolution priors.
\newblock In \emph{ACM MM}, 2022.

\bibitem[Chen et~al.(2024)Chen, Yu, Ge, Yao, Xie, Wu, Wang, Kwok, Luo, Lu,
  et~al.]{pixart}
Junsong Chen, Jincheng Yu, Chongjian Ge, Lewei Yao, Enze Xie, Yue Wu, Zhongdao
  Wang, James Kwok, Ping Luo, Huchuan Lu, et~al.
\newblock Pixart-$\alpha$: Fast training of diffusion transformer for
  photorealistic text-to-image synthesis.
\newblock In \emph{ICLR}, 2024.

\bibitem[Gu et~al.(2019{\natexlab{a}})Gu, Lu, Zuo, and Dong]{IKC}
Jinjin Gu, Hannan Lu, Wangmeng Zuo, and Chao Dong.
\newblock Blind super-resolution with iterative kernel correction.
\newblock In \emph{CVPR}, 2019{\natexlab{a}}.

\bibitem[Gu et~al.(2019{\natexlab{b}})Gu, Lugmayr, Danelljan, Fritsche, Lamour,
  and Timofte]{div8k}
Shuhang Gu, Andreas Lugmayr, Martin Danelljan, Manuel Fritsche, Julien Lamour,
  and Radu Timofte.
\newblock Div8k: Diverse 8k resolution image dataset.
\newblock In \emph{ICCV Workshops}, 2019{\natexlab{b}}.

\bibitem[Ho and Salimans(2022)]{classifier}
Jonathan Ho and Tim Salimans.
\newblock Classifier-free diffusion guidance.
\newblock \emph{arXiv preprint arXiv:2207.12598}, 2022.

\bibitem[Ho et~al.(2020)Ho, Jain, and Abbeel]{DDPM}
Jonathan Ho, Ajay Jain, and Pieter Abbeel.
\newblock Denoising diffusion probabilistic models.
\newblock In \emph{NeurIPS}, 2020.

\bibitem[Hu et~al.(2024)Hu, Gao, Zhang, Sun, Zhang, and Bo]{Animate}
Li Hu, Xin Gao, Peng Zhang, Ke Sun, Bang Zhang, and Liefeng Bo.
\newblock Animate anyone: Consistent and controllable image-to-video synthesis
  for character animation.
\newblock In \emph{CVPR}, 2024.

\bibitem[Huang et~al.(2020)Huang, Li, Wang, Tan, et~al.]{unfoldingSR}
Yan Huang, Shang Li, Liang Wang, Tieniu Tan, et~al.
\newblock Unfolding the alternating optimization for blind super resolution.
\newblock In \emph{NeurIPS}, 2020.

\bibitem[Karras et~al.(2019)Karras, Laine, and Aila]{FFHQ}
Tero Karras, Samuli Laine, and Timo Aila.
\newblock A style-based generator architecture for generative adversarial
  networks.
\newblock In \emph{CVPR}, 2019.

\bibitem[Karras et~al.(2022)Karras, Aittala, Aila, and Laine]{elucidating}
Tero Karras, Miika Aittala, Timo Aila, and Samuli Laine.
\newblock Elucidating the design space of diffusion-based generative models.
\newblock In \emph{NeurIPS}, 2022.

\bibitem[Ke et~al.(2021)Ke, Wang, Wang, Milanfar, and Yang]{MUSIQ}
Junjie Ke, Qifei Wang, Yilin Wang, Peyman Milanfar, and Feng Yang.
\newblock Musiq: Multi-scale image quality transformer.
\newblock In \emph{CVPR}, 2021.

\bibitem[Kingma(2014)]{Autoencoder}
Diederik~P Kingma.
\newblock Auto-encoding variational bayes.
\newblock In \emph{ICLR}, 2014.

\bibitem[Li et~al.(2022)Li, Liu, Guo, Yin, Jiang, Du, Du, Zhu, Lai, Hu,
  et~al.]{PPOCR}
Chenxia Li, Weiwei Liu, Ruoyu Guo, Xiaoting Yin, Kaitao Jiang, Yongkun Du,
  Yuning Du, Lingfeng Zhu, Baohua Lai, Xiaoguang Hu, et~al.
\newblock Pp-ocrv3: More attempts for the improvement of ultra lightweight ocr
  system.
\newblock \emph{arXiv preprint arXiv:2206.03001}, 2022.

\bibitem[Li et~al.(2023)Li, Zhang, Liang, Cao, Liu, Gong, Zhang, Tang, Liu,
  Demandolx, et~al.]{LSDIR}
Yawei Li, Kai Zhang, Jingyun Liang, Jiezhang Cao, Ce Liu, Rui Gong, Yulun
  Zhang, Hao Tang, Yun Liu, Denis Demandolx, et~al.
\newblock Lsdir: A large scale dataset for image restoration.
\newblock In \emph{CVPR}, 2023.

\bibitem[Liang et~al.(2022{\natexlab{a}})Liang, Zeng, and Zhang]{DASR}
Jie Liang, Hui Zeng, and Lei Zhang.
\newblock Efficient and degradation-adaptive network for real-world image
  super-resolution.
\newblock In \emph{ECCV}, 2022{\natexlab{a}}.

\bibitem[Liang et~al.(2022{\natexlab{b}})Liang, Zeng, and Zhang]{LDL}
Jie Liang, Hui Zeng, and Lei Zhang.
\newblock Details or artifacts: A locally discriminative learning approach to
  realistic image super-resolution.
\newblock In \emph{CVPR}, 2022{\natexlab{b}}.

\bibitem[Lim et~al.(2017)Lim, Son, Kim, Nah, and Mu~Lee]{Flicker2k}
Bee Lim, Sanghyun Son, Heewon Kim, Seungjun Nah, and Kyoung Mu~Lee.
\newblock Enhanced deep residual networks for single image super-resolution.
\newblock In \emph{CVPR Workshops}, 2017.

\bibitem[Lin et~al.(2023)Lin, He, Chen, Lyu, Dai, Yu, Ouyang, Qiao, and
  Dong]{DiffBIR}
Xinqi Lin, Jingwen He, Ziyan Chen, Zhaoyang Lyu, Bo Dai, Fanghua Yu, Wanli
  Ouyang, Yu Qiao, and Chao Dong.
\newblock Diffbir: Towards blind image restoration with generative diffusion
  prior.
\newblock \emph{arXiv preprint arXiv:2308.15070}, 2023.

\bibitem[Liu et~al.(2024)Liu, Li, Wu, and Lee]{LLAVA}
Haotian Liu, Chunyuan Li, Qingyang Wu, and Yong~Jae Lee.
\newblock Visual instruction tuning.
\newblock In \emph{NeurIPS}, 2024.

\bibitem[Loshchilov and Hutter(2017)]{cosine}
Ilya Loshchilov and Frank Hutter.
\newblock Sgdr: Stochastic gradient descent with warm restarts.
\newblock In \emph{ICLR}, 2017.

\bibitem[Loshchilov and Hutter(2019)]{Adamw}
Ilya Loshchilov and Frank Hutter.
\newblock Decoupled weight decay regularization.
\newblock In \emph{ICLR}, 2019.

\bibitem[Mou et~al.(2024)Mou, Wang, Xie, Wu, Zhang, Qi, and Shan]{T2I_adapter}
Chong Mou, Xintao Wang, Liangbin Xie, Yanze Wu, Jian Zhang, Zhongang Qi, and
  Ying Shan.
\newblock T2i-adapter: Learning adapters to dig out more controllable ability
  for text-to-image diffusion models.
\newblock In \emph{AAAI}, 2024.

\bibitem[Podell et~al.(2024)Podell, English, Lacey, Blattmann, Dockhorn,
  M{\"u}ller, Penna, and Rombach]{SDXL}
Dustin Podell, Zion English, Kyle Lacey, Andreas Blattmann, Tim Dockhorn, Jonas
  M{\"u}ller, Joe Penna, and Robin Rombach.
\newblock Sdxl: Improving latent diffusion models for high-resolution image
  synthesis.
\newblock In \emph{ICLR}, 2024.

\bibitem[Qu et~al.(2024)Qu, Yuan, Zhao, Xie, Hao, Sun, and Zhou]{XPSR}
Yunpeng Qu, Kun Yuan, Kai Zhao, Qizhi Xie, Jinhua Hao, Ming Sun, and Chao Zhou.
\newblock Xpsr: Cross-modal priors for diffusion-based image super-resolution.
\newblock In \emph{ECCV}, 2024.

\bibitem[Radford et~al.(2021)Radford, Kim, Hallacy, Ramesh, Goh, Agarwal,
  Sastry, Askell, Mishkin, Clark, et~al.]{CLIP}
Alec Radford, Jong~Wook Kim, Chris Hallacy, Aditya Ramesh, Gabriel Goh,
  Sandhini Agarwal, Girish Sastry, Amanda Askell, Pamela Mishkin, Jack Clark,
  et~al.
\newblock Learning transferable visual models from natural language
  supervision.
\newblock In \emph{ICML}, 2021.

\bibitem[Ramesh et~al.(2022)Ramesh, Dhariwal, Nichol, Chu, and Chen]{HCL}
Aditya Ramesh, Prafulla Dhariwal, Alex Nichol, Casey Chu, and Mark Chen.
\newblock Hierarchical text-conditional image generation with clip latents.
\newblock \emph{arXiv preprint arXiv:2204.06125}, 2022.

\bibitem[Rombach et~al.(2022{\natexlab{a}})Rombach, Blattmann, Lorenz, Esser,
  and Ommer]{HLDM}
Robin Rombach, Andreas Blattmann, Dominik Lorenz, Patrick Esser, and Bj{\"o}rn
  Ommer.
\newblock High-resolution image synthesis with latent diffusion models.
\newblock In \emph{CVPR}, 2022{\natexlab{a}}.

\bibitem[Rombach et~al.(2022{\natexlab{b}})Rombach, Blattmann, Lorenz, Esser,
  and Ommer]{SD}
Robin Rombach, Andreas Blattmann, Dominik Lorenz, Patrick Esser, and Bj{\"o}rn
  Ommer.
\newblock High-resolution image synthesis with latent diffusion models.
\newblock In \emph{CVPR}, 2022{\natexlab{b}}.

\bibitem[Saharia et~al.(2022)Saharia, Chan, Saxena, Li, Whang, Denton,
  Ghasemipour, Gontijo~Lopes, Karagol~Ayan, Salimans, et~al.]{imagen}
Chitwan Saharia, William Chan, Saurabh Saxena, Lala Li, Jay Whang, Emily~L
  Denton, Kamyar Ghasemipour, Raphael Gontijo~Lopes, Burcu Karagol~Ayan, Tim
  Salimans, et~al.
\newblock Photorealistic text-to-image diffusion models with deep language
  understanding.
\newblock In \emph{NeurIPS}, 2022.

\bibitem[Tang et~al.(2022)Tang, Liu, Pandey, Jiang, Yang, Kumar, Stenetorp,
  Lin, and Ture]{DAAM}
Raphael Tang, Linqing Liu, Akshat Pandey, Zhiying Jiang, Gefei Yang, Karun
  Kumar, Pontus Stenetorp, Jimmy Lin, and Ferhan Ture.
\newblock What the daam: Interpreting stable diffusion using cross attention.
\newblock \emph{arXiv preprint arXiv:2210.04885}, 2022.

\bibitem[Tian et~al.(2024)Tian, Wang, Zhang, and Bo]{EMO}
Linrui Tian, Qi Wang, Bang Zhang, and Liefeng Bo.
\newblock Emo: Emote portrait alive - generating expressive portrait videos
  with audio2video diffusion model under weak conditions.
\newblock In \emph{ECCV}, 2024.

\bibitem[Tom et~al.(2024)Tom, Mathew, Mondal, Karatzas, Jawahar, and
  Weinman]{ICDAR}
George Tom, Minesh Mathew, Ajoy Mondal, Dimosthenis Karatzas, C.~V. Jawahar,
  and Jerod Weinman.
\newblock Icdar2024 challenge on occluded roadtext.
\newblock In \emph{ICDAR2024 Workshops}, 2024.

\bibitem[Vaswani(2017)]{Transformer}
A Vaswani.
\newblock Attention is all you need.
\newblock In \emph{NeurIPS}, 2017.

\bibitem[Wang et~al.(2023)Wang, Chan, and Loy]{CLIPIQA}
Jianyi Wang, Kelvin~CK Chan, and Chen~Change Loy.
\newblock Exploring clip for assessing the look and feel of images.
\newblock In \emph{AAAI}, 2023.

\bibitem[Wang et~al.(2024)Wang, Yue, Zhou, Chan, and Loy]{StableSR}
Jianyi Wang, Zongsheng Yue, Shangchen Zhou, Kelvin~CK Chan, and Chen~Change
  Loy.
\newblock Exploiting diffusion prior for real-world image super-resolution.
\newblock \emph{IJCV}, 2024.

\bibitem[Wang et~al.(2021{\natexlab{a}})Wang, Li, Zhang, and Shan]{GFPGAN}
Xintao Wang, Yu Li, Honglun Zhang, and Ying Shan.
\newblock Towards real-world blind face restoration with generative facial
  prior.
\newblock In \emph{CVPR}, 2021{\natexlab{a}}.

\bibitem[Wang et~al.(2021{\natexlab{b}})Wang, Xie, Dong, and Shan]{Real-ESRGAN}
Xintao Wang, Liangbin Xie, Chao Dong, and Ying Shan.
\newblock Real-esrgan: Training real-world blind super-resolution with pure
  synthetic data.
\newblock In \emph{CVPR}, 2021{\natexlab{b}}.

\bibitem[Wu et~al.(2024)Wu, Yang, Sun, Zhang, Li, and Zhang]{SeeSR}
Rongyuan Wu, Tao Yang, Lingchen Sun, Zhengqiang Zhang, Shuai Li, and Lei Zhang.
\newblock Seesr: Towards semantics-aware real-world image super-resolution.
\newblock In \emph{CVPR}, 2024.

\bibitem[Xie et~al.(2023)Xie, Wang, Chen, Li, Shan, Zhou, and Dong]{Desra}
Liangbin Xie, Xintao Wang, Xiangyu Chen, Gen Li, Ying Shan, Jiantao Zhou, and
  Chao Dong.
\newblock Desra: detect and delete the artifacts of gan-based real-world
  super-resolution models.
\newblock In \emph{ICML}, 2023.

\bibitem[Yang et~al.(2024)Yang, Wu, Ren, Xie, and Zhang]{PASD}
Tao Yang, Rongyuan Wu, Peiran Ren, Xuansong Xie, and Lei Zhang.
\newblock Pixel-aware stable diffusion for realistic image super-resolution and
  personalized stylization.
\newblock In \emph{ECCV}, 2024.

\bibitem[Yu et~al.(2024)Yu, Gu, Li, Hu, Kong, Wang, He, Qiao, and Dong]{SUPIR}
Fanghua Yu, Jinjin Gu, Zheyuan Li, Jinfan Hu, Xiangtao Kong, Xintao Wang,
  Jingwen He, Yu Qiao, and Chao Dong.
\newblock Scaling up to excellence: Practicing model scaling for
  photo-realistic image restoration in the wild.
\newblock In \emph{CVPR}, 2024.

\bibitem[Zhang et~al.(2018{\natexlab{a}})Zhang, Zuo, and Zhang]{SRMD}
Kai Zhang, Wangmeng Zuo, and Lei Zhang.
\newblock Learning a single convolutional super-resolution network for multiple
  degradations.
\newblock In \emph{CVPR}, 2018{\natexlab{a}}.

\bibitem[Zhang et~al.(2021)Zhang, Liang, Van~Gool, and Timofte]{BSRGAN}
Kai Zhang, Jingyun Liang, Luc Van~Gool, and Radu Timofte.
\newblock Designing a practical degradation model for deep blind image
  super-resolution.
\newblock In \emph{CVPR}, 2021.

\bibitem[Zhang et~al.(2023)Zhang, Rao, and Agrawala]{controlnet}
Lvmin Zhang, Anyi Rao, and Maneesh Agrawala.
\newblock Adding conditional control to text-to-image diffusion models.
\newblock In \emph{ICCV}, 2023.

\bibitem[Zhang et~al.(2018{\natexlab{b}})Zhang, Isola, Efros, Shechtman, and
  Wang]{Lpips}
Richard Zhang, Phillip Isola, Alexei~A Efros, Eli Shechtman, and Oliver Wang.
\newblock The unreasonable effectiveness of deep features as a perceptual
  metric.
\newblock In \emph{CVPR}, 2018{\natexlab{b}}.

\bibitem[Zhou et~al.(2022)Zhou, Chan, Li, and Loy]{Codeformer}
Shangchen Zhou, Kelvin Chan, Chongyi Li, and Chen~Change Loy.
\newblock Towards robust blind face restoration with codebook lookup
  transformer.
\newblock In \emph{NeurIPS}, 2022.

\end{thebibliography}
